\documentclass[pdflatex,sn-mathphys]{sn-jnl}

\jyear{2022}%

\theoremstyle{thmstyleone}%
\theoremstyle{thmstyletwo}%

\theoremstyle{thmstylethree}%

\begin{document}

\title[Composite Anomaly Detection for Imbalanced Industrial Dataset]{Composite Score for Anomaly Detection in Imbalanced Real-World Industrial Dataset}

\author*[1]{\fnm{Arnaud} \sur{Bougaham}}\email{arnaud.bougaham@unamur.be}

\author[1]{\fnm{Mohammed} \sur{El Adoui}}\email{mohammed.eladoui@unamur.be}

\author[2]{\fnm{Isabelle} \sur{Linden}}\email{isabelle.linden@unamur.be}

\author[1]{\fnm{Beno\^it} \sur{Fr\'enay}}\email{benoit.frenay@unamur.be}

\affil*[1]{\orgdiv{Faculty of Computer Science, NaDI Institute}, \\\orgname{University of Namur}, \orgaddress{\street{Rue Grandgagnage 21}, \city{Namur}, \postcode{5000}, \country{Belgium}}}

\affil[2]{\orgdiv{Department of Management Sciences, NaDI Institute}, \\\orgname{University of Namur}, \orgaddress{\street{Rempart de la Vierge 8}, \city{Namur}, \postcode{5000}, \country{Belgium}}}

\abstract{In recent years, the industrial sector has evolved towards its fourth revolution. The quality control domain is particularly interested in advanced machine learning for computer vision anomaly detection. Nevertheless, several challenges have to be faced, including imbalanced datasets, the image complexity, and the zero-false-negative (ZFN) constraint to guarantee the high-quality requirement. This paper illustrates a use case for an industrial partner, where Printed Circuit Board Assembly (PCBA) images are first reconstructed with a Vector Quantized Generative Adversarial Network (\textit{VQGAN}) trained on normal products. Then, several multi-level metrics are extracted on a few normal and abnormal images, highlighting anomalies through reconstruction differences. Finally, a classifier is trained to build a composite anomaly score thanks to the metrics extracted. This three-step approach is performed on the public MVTec-AD datasets and on the partner PCBA dataset, where it achieves a regular accuracy of 95.69\% and 87.93\% under the ZFN constraint.}

\keywords{Imbalanced Learning, Industry 4.0, Anomaly Detection, High Resolution Images, Zero False Negative, Computer Vision, Real-World, PCBA}



\maketitle

\section{Introduction}\label{sec1}
Anomaly detection is a ubiquitous concern for industries ensuring robust manufacturing quality control. Operation sites considering this challenge need, for instance, high-resolution images of the product being manufactured so that an anomaly detection method can be executed. Usually, an automatic inspection process is devoted to perform this task. It takes several images of the same product with different view angles and lighting conditions. Then, it asks an operator to confirm or infirm a pseudo-error if a doubt on the product quality is raised. Negative samples are anomaly-free images, unlike positive ones where a defect is observed on the product image. Severe test limits are necessary to avoid missed detections (false negatives or type II errors), depending on the quality strategy. This way, the defect is early detected, and the abnormal product is not propagated towards the next production processes. The drawback of this strategy is a high rate of false alarms (false positives or type I errors) due to the product image complexity. 

In practice, it causes fault investigation losses and product retests. Moreover, the human operator gets overflowed by the inspection process, often incorrectly considering negative products as positive. The credibility of this anomaly detection process is therefore reduced. This yields in some human misjudgments, classifying positives, well detected by the inspection process, as false negatives, because of the habit of invalidating the process pseudo-errors. Should this happens, the product is stopped later on the production line, but the repair or scrap costs are greater. If it can be repaired, the time needed to access the defect area or component increases due to all the parts composing the product. If the product has to be scrapped, the processes and the workers time needed to manufacture it is lost, as well as the components placed after this inspection process. Consequently, valuable time is wasted, repair costs are increased, and quality risks are taken. A key challenge is therefore to reduce the false alarms, keeping the requirement to avoid any missed detection.\\

This work is carried out with an industrial partner specialized in Printed Circuit Board Assembly (PCBA) for the automotive industry. These PCBAs are devoted to provide automatic car speed boxes after being sealed in a case, forming an Automatic Transmission Electronic Control Unit (ATECU). The production line is composed of 3 distinct blocks: the placement and soldering of the electronic components on the blank circuit, the connector and case assembly steps, and the final product test stage. All the processes are placed inline to manufacture the product step by step, with quality inspections alongside. 

It is commonly agreed that the earlier a defect is detected, the earlier it can be contained. Based on this statement, the scope of this work is focused on the optical inspection, the first visual test. Its main role is to take images of 100\% of the products to estimate the quality at the very first stages of the production line, where electronic components are placed and soldered. This critical process ensures an early reaction if needed, making it possible to countermeasure the eventual issues and avoid propagating the defect downstream. A telemetry-based image processing algorithm is currently in charge of comparing the image being treated with a golden sample image, to guarantee the anomaly detection. 

\begin{figure}[h]
    \centering
    {\includegraphics[width=0.28\linewidth]{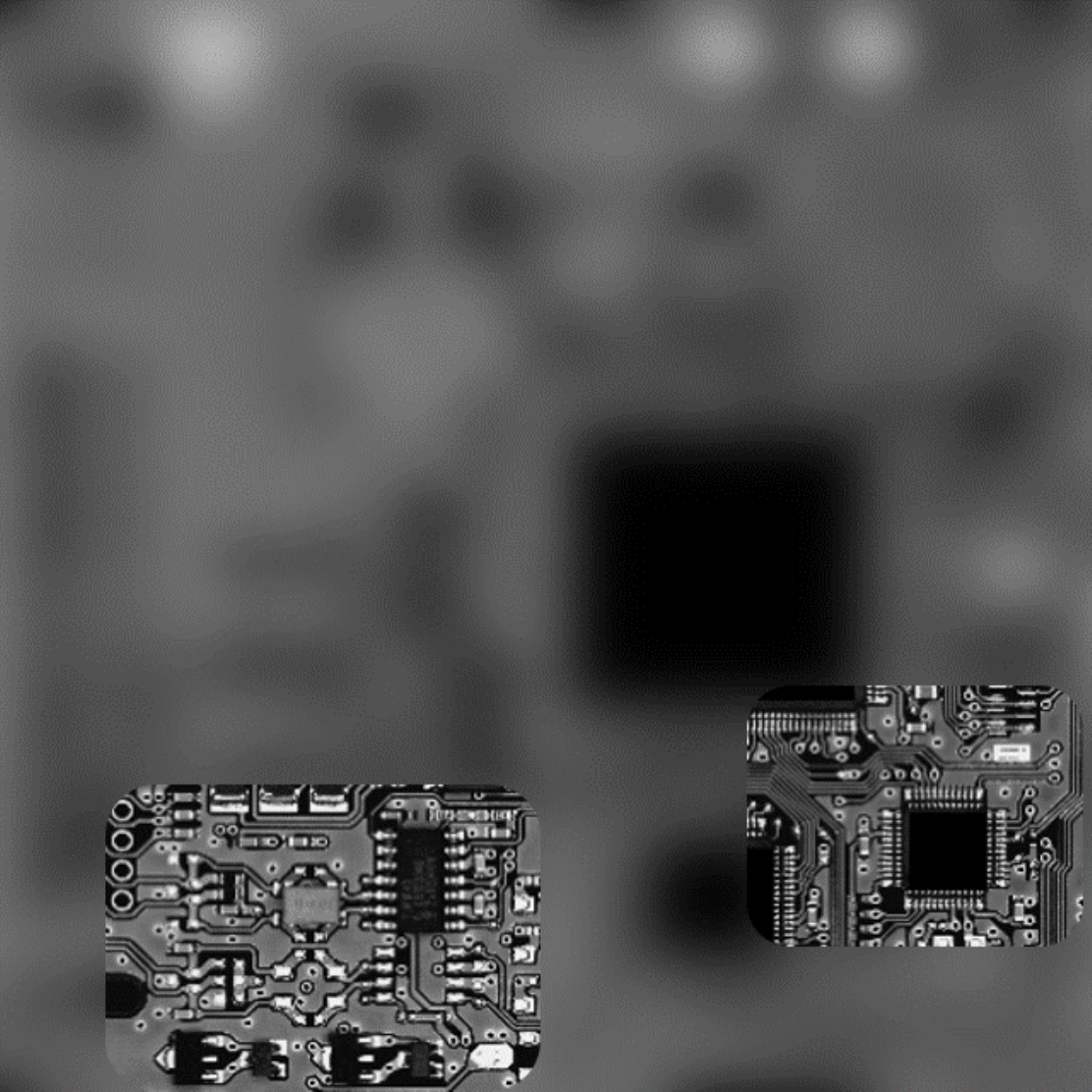}}
    \qquad
    {\includegraphics[width=0.28\linewidth]{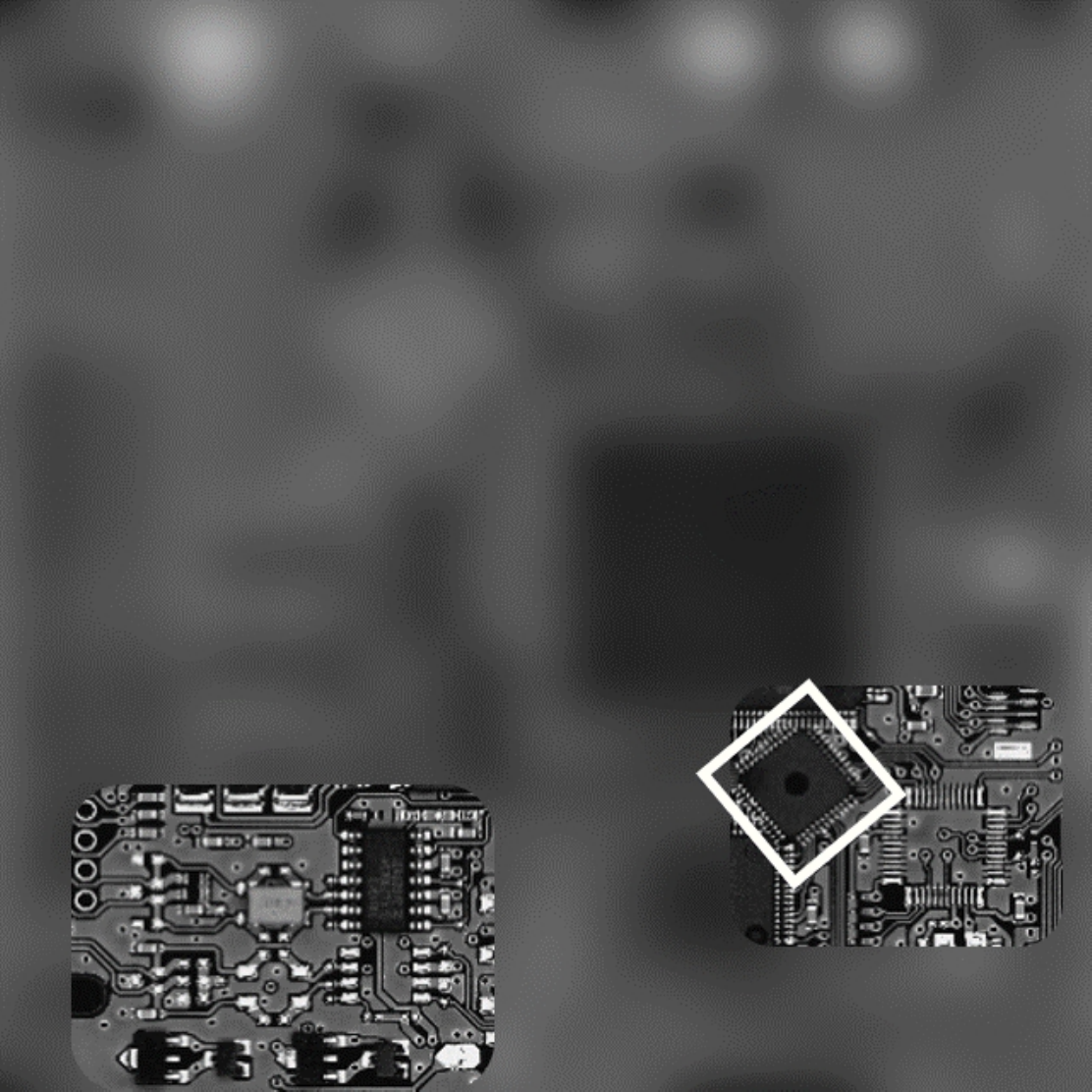}}
    \qquad
    {\includegraphics[width=0.28\linewidth]{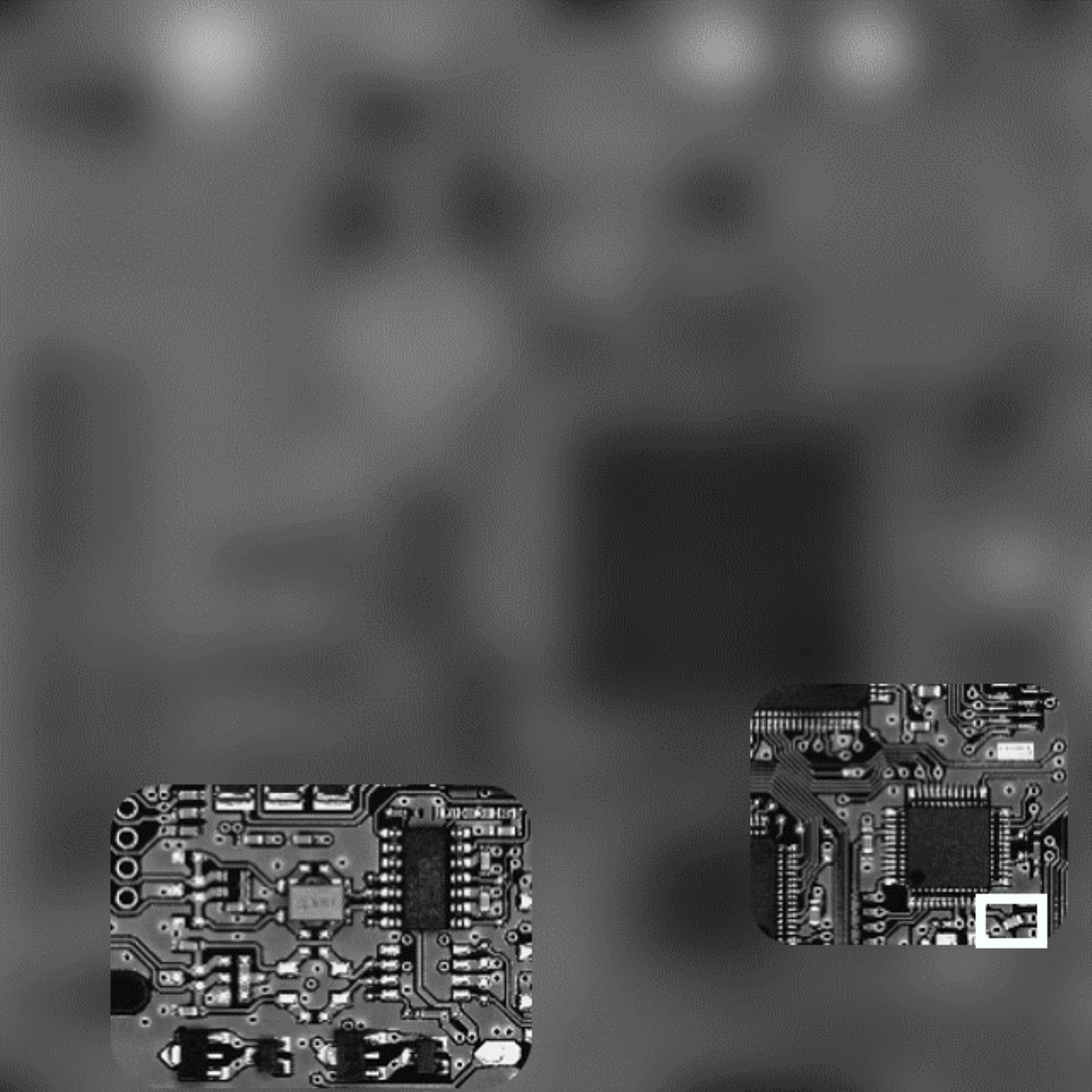}}
  \caption{The left image presents a normal PCBA. On the other ones, the white frames surround a large anomaly (middle image) and a small one (right image). Some parts of the images have been anonymized (material under intellectual property).}
  \label{fig:instance_example}
\end{figure}

Some image examples are shown in Figure~\ref{fig:instance_example}\footnote{Some parts of the images have been blurred to guarantee the intellectual property of our industrial partner. The arguments described also apply to the hidden parts, where information can be extrapolated.} (presented in \cite{bougaham2021ganodip}), where one can figure out the details in terms of objects, reflections, or texture.
The severe test limits imply a large number of false positives raised by the algorithm, which slows down the throughput, and requires tedious labor for human operators to make the final judgment. A standard acceptable operator misjudgment rate in these conditions is 2\% of the production \cite{down2010measurement} (true positives detected by the inspection process but incorrectly judged as negatives by the human operator), which is far higher than the actual rate of our industrial partner. However, this low human variability means that multiple efforts and training costs are spent on reducing as possible these false negatives. The initial claim is an average time loss of $\pm 8\text{s/PCBA}$ due to the algorithm false-positive rate, and around 83 parts per million abnormal PCBAs misjudged as normal by the operator, due to process credibility reduction.\\

Recently, deep learning methods have attracted much interest in the context of Industry 4.0, as they can help alleviate the problem of type I and type II errors \cite{xia2022gan}. Thanks to a vast number of images, such an advanced method can be performed to state whether a product is standard or not. Therefore, these techniques can supplement or even replace traditional anomaly detection systems. However, due to the imbalanced nature of the datasets at hand, it is challenging to design a regular binary classifier. Indeed, the extreme rarity of anomalies yields many more images with normal products and a few images with anomalies. This lack of minority-class images leads industries to deal with representation learning techniques, suitable for extracting an insightful feature representation of the majority class. In a first step, a model learns the normal data distribution, and, in a second step, this model reconstructs a new input image under test, based on this normal representation model. Finally, a distance is computed between the input and the reconstructed images. This anomaly score states how different both images are, and a threshold defines the normality of the input image, under the assumption that the model will recover the eventual abnormal set of pixels.

\begin{figure}[h]
    \centering
    \includegraphics[width=0.9\linewidth]{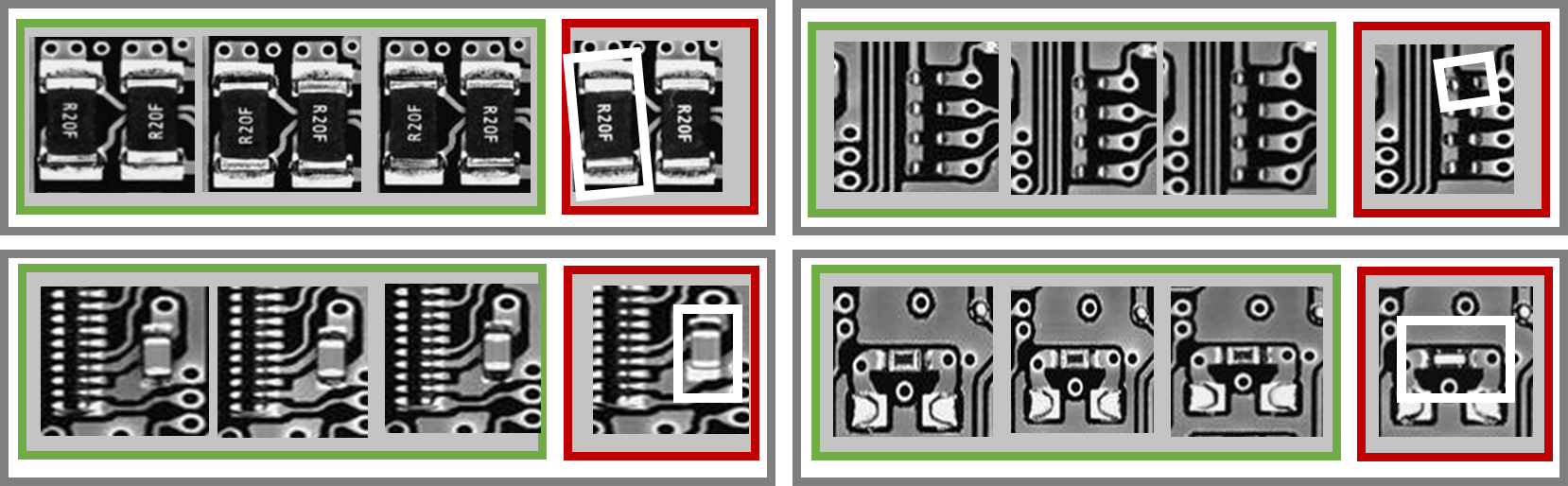}
    \caption{(Color online) Four blocks of zoomed areas ($\approx X10$) for 4 different PCBA images. For each block, the left green frame (3 first columns) shows normal variations unlike the red frame one (last columns), where a small defect is observed. The challenge is to discriminate normal and abnormal variations. These very small areas represent around $1\text{cm} \times 1\text{cm}$ over the $10\text{cm} \times 10\text{cm}$ surface of the entire product.}
    \label{fig:Variations}
\end{figure}

The reconstruction quality is a difficult task for complex images, but the difference between normal and abnormal variability is another substantial difficulty. For images with many details, as is the case for the PCBAs, a macro view does not reveal high variability. Nevertheless, in detail, the PCBAs are showing many differences. Therefore, one of our challenges is to distinguish between a small defect and a slight normal variation. Figure~\ref{fig:Variations} shows such normal and abnormal variations in 4 different zoomed areas of the images, where one can appreciate the very small difference in component shift for both cases.

Based on the Vector Quantized Generative Adversarial Network (\textit{VQGAN}) \cite{esser2021taming} and the Generative Adversarial Network Anomaly Detection Through Intermediate Patches (\textit{GanoDIP}) \cite{bougaham2021ganodip} works, a methodology called \textit{\textbf{VQGanoDIP}} is proposed. It is aimed to tackle the imbalance and complexity dimension of the real-world industrial PCBA dataset, composed of high-resolution images with a fine distinction between normal and abnormal variations. The main contribution is (i) to associate these works to get the best representation possible of the majority class, in addition (ii) to develop techniques (such as weighting normal variations or multi-level distances collection) to localize estimated defect areas, and (iii) to compute a composite anomaly score that characterizes them through a binary classifier. The objective is to reduce the false-positive rate while enforcing the zero-false-negative (ZFN) rate requirement.

First, a literature review on industrial anomaly detection and image synthesis is proposed in Section~\ref{Related_Works}. Afterward, Section~\ref{VQGanoDIP_method} details the \textit{VQGanoDIP} methodology, stating how the three-step (reconstruction, metrics extraction, and normal/abnormal classification) can achieve the objectives. Finally, the method performance is qualitatively and quantitatively reported and discussed in Section~\ref{Evaluation}, on multiple datasets.

\section{Related Works}\label{Related_Works}

Computer vision anomaly detection presents a significant interest within multiple domains. Nowadays, several studies are dedicated to summarize these methodologies. Xia et al. \cite{xia2022gan} reviewed various existing methods applying deep learning algorithms such as CNNs and GANs applied to anomaly detection. According to this study, these methods often depend on large training samples. Therefore, data imbalance is one of the main application limitations. They state that GANs are one of the best solutions proposed in the literature to deal with it, by learning representation features of the majority class, in an unsupervised manner. Akçay et al. \cite{akccay2019skip} introduced an approach using unsupervised anomaly detection within a GAN training scheme. This approach is based on an autoencoder with skip-connections, terminated by a GAN discriminator that provides effective training for the normal class. However, they suggested to apply their method on higher resolution images as future work. This identified limitation is indeed an obstacle when small defects detection is a strong requirement, which is our case. Schlegl et al. \cite{schlegl2017unsupervised} explored the encoded latent vector thanks to a GAN generator learning the data distribution. First, the authors trained a generator and a discriminator using images without anomaly. Then, the pre-trained generator and discriminator are frozen, and a latent vector mapping is performed. Despite the high performance reported, its computational complexity remains expensive. In addition, the authors limited their experiments to low-resolution images and applied them to a unique type of images (retina optical coherence tomography scan), unlike the approach we introduced where the genericity dimension is considered.

In our previous work \cite{bougaham2021ganodip}, we proposed to use intermediate patches for the inference step after a Wasserstein GAN (WGAN) training procedure. Our objective was to make anomaly detection possible on real-world industrial Printed Circuit Board Assembly (PCBA) images. This approach showed that our previous technique can be used to support current industrial image processing algorithms and avoid wasting time for industries using manual techniques. However, real-world implementation is still challenging, due to the high diversity of defects possible in a PCBA, particularly the very small ones, undetectable below the megapixel resolution. Van Den Oord et al. \cite{van2017neural} incorporated the concept of vector quantization (\textit{VQ-VAE}) in order to learn a discrete latent representation. Following their methodology, the model is able to generate expressive images and speech data. Still the image resolution considered in this work is not sufficient to be considered for our high-resolution constraint. Razavi et al. \cite{razavi2019generating} improved the Vector Quantized Variational AutoEncoder (\textit{VQ-VAE2}) models for large-scale image generation. They enhanced the auto-regressive priors used in their architecture to produce synthetic samples of higher coherence. One of their main contributions was to keep the encoder-decoder architecture simple and lightweight. Regardless of the performance demonstrated by the VAE architectures introduced by these two studies, Esser et al. \cite{esser2021taming} showed that the \textit{VQ-VAE} methods produced reconstructions yielding blurred details, being an issue to reconstruct our PCBA images with sufficient fidelity. They addressed this limitation by synthesizing realistic detailed high-resolution images with a Vector Quantized Generative Adversarial Network (\textit{VQGAN}). Their approach, based on \textit{VQ-VAE}, consisted of representing the images as a composition of coherent and rich details, adding a GAN discriminator to improve the images realness, and considering a perceptual loss.

In addition to the anomaly detection and image synthesis problems, our business specificity requires to guarantee that no defect can be missed by the algorithm. Some studies consider the classification as an optimization or a cost-sensitive problem (\cite{sangalli2021constrained},\cite{krawczyk2014cost}), in order to prioritize the false positive misclassifications instead of the false negative ones. Although these works consider the constraint in an end-to-end manner, the missed detections are minimized without any guarantee on their absence. Roth et al. \cite{roth2021towards} proposed to adjust a 100\% recall threshold for predictions of a patch-features encoding anomaly detection method, for industrial public dataset images. Their method exploits a threshold that guarantees no false negatives, but operates at low resolution, being an issue for our PCBA anomaly detection task.

Considering the above studies using quantized autoencoder with GAN methodologies and showing promising results within the field of anomaly detection, we propose a new approach exploiting adversarial quantized auto-encoders to reconstruct an input image, collect metrics from this reconstruction, train a binary classifier on this metrics dataset, for high-resolution and challenging real-world images. Such a method aims to discriminate between anomalies that are not necessarily clear and patterned compared to the normal variation, and to guarantee that no missed detection is possible (frequent requirement for the industrial or medical applications).

\section{VQGanoDIP: VQGAN Anomaly Detection through Intermediate Patches}\label{VQGanoDIP_method}
This section details the proposed \textit{VQGanoDIP} (\textit{VQGAN} + \textit{GanoDIP}) methodology, designed to localize and quantify abnormal areas in the PCBA and the MVTEC-AD $1024 \times 1024$ images \cite{bergmann2019mvtec}, a set of different high-resolution industrial images composed of products with and without anomalies. The first step is based on \textit{VQGAN} \cite{esser2021taming}, in particular, the reconstruction part. Eventual anomalies are expected to be recovered on a set of images of the two classes, thanks to the model previously trained on normal the majority class, spotting the differences between the input and the reconstruction. Based on these differences, a highlighting technique inspired by our previous work \cite{bougaham2021ganodip} is performed, where a patching method localizes anomalies in a reduced set of areas. Then, the differences are quantified with several metrics, image-wise, patch-wise, and pixel-wise. Finally, these collected metrics on normal and abnormal images are used to train a binary classifier model that compute a composite anomaly score qualifying the product quality. To determine the product quality, a threshold is adjusted to avoid missing any true positive. 

\subsection{First step: Image Reconstruction \& Anomaly Localization}\label{subsec3}
The first step of the \textit{VQGanoDIP} methodology is the image reconstruction and the most abnormal sets of pixels localization.

\subsubsection{VQGAN Reconstruction}\label{subsubsecVQGAN}
The \textit{VQGAN} method has been selected for its ability to efficiently learn a data representation and synthesize small details in an information-rich image. The specificity of our PCBA dataset lies in the fact that the images are similar in a global manner, with no significant variation in the component placement, the circuit color, or the solder pads. However, it offers a lot of small local variations due to the placement and solder process windows. The CNNs inductive bias that encourages the local interactions coming from this method allows dealing with these small variations and can follow the data distribution locally and globally. Moreover, its ``context-rich vocabulary learning of the image constituents'' \cite{esser2021taming} reduces the practical computational resources and allows generation in the megapixel regime, which makes it possible to work in a high-resolution space and thus capture very small defects.

\begin{figure}[htbp]
    \centering
    \includegraphics[width=0.8\linewidth]{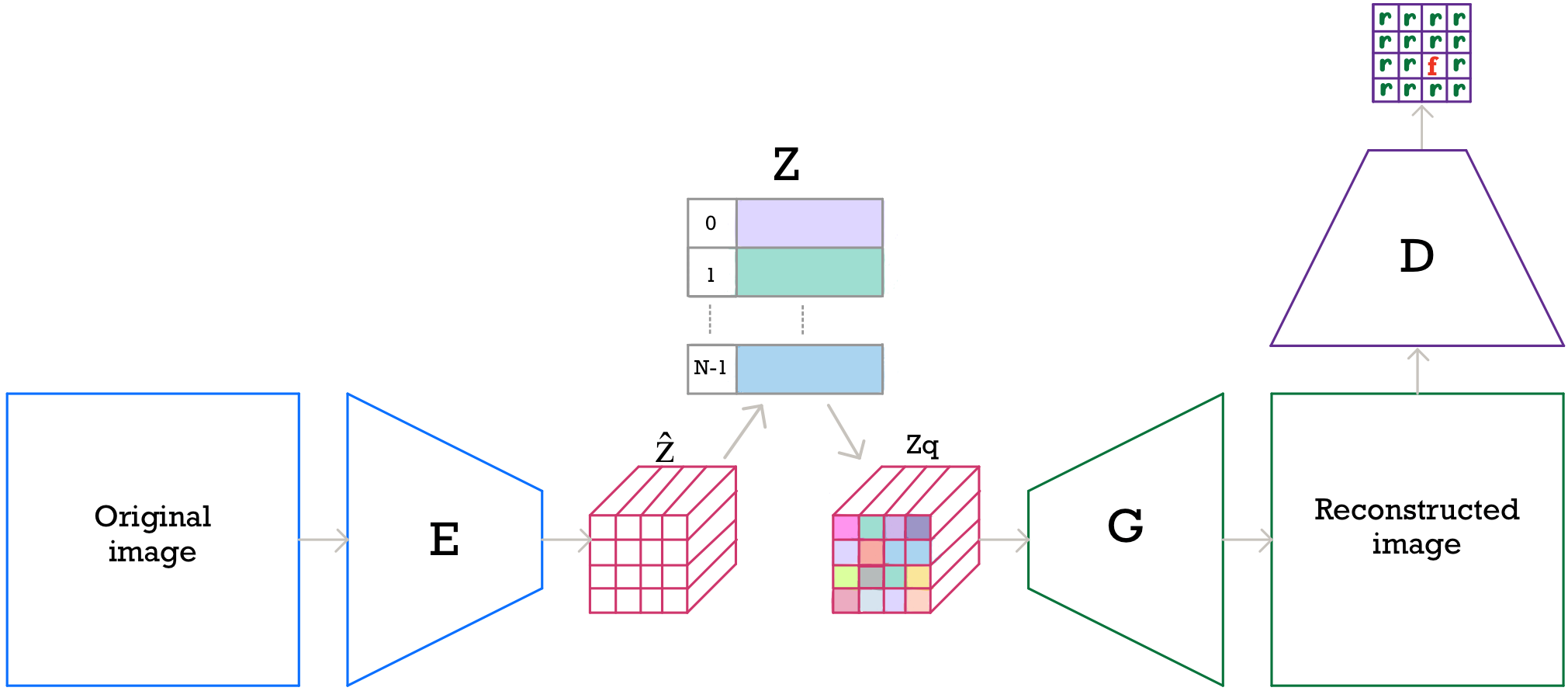}
    \caption{(Color online) Figure inspired from \cite{esser2021taming} presenting the training strategy of \textit{VQGAN}, which is used as the first step of the proposed \textit{VQGanoDIP} methodology.}
    \label{fig:VQGAN}
\end{figure}

\textit{VQGAN} is a vector quantized autoencoder model augmented with a GAN. On top of this method, image synthesis is achievable thanks to a transformer architecture (out of our scope since the objective is only the reconstruction). Figure~\ref{fig:VQGAN} shows the framework of the \textit{VQGAN} reconstruction model. 

The architecture is composed of 3 stages.
First, the central part of the framework is an autoencoder. The encoder $E$ learns a mapping function to transform the high-dimensional original image into a low-dimensional latent representation. Afterward, a reconstructed high-dimensional image is decoded from the latent representation, thanks to the latent-image space mapping, which is the decoder (or generator $G$).

The second stage is the vector quantized (VQ) part of this autoencoder. It adds the advantage of transforming the learned latent representation $\hat{\text{z}}$ into a quantized representation $z_q$, instead of a continuous one, yielding many possible values to be decoded (difficult to learn). Therefore the model is able to focus on a restricted number of latent vectors, which significantly helps model convergence and avoids mode collapse (identified in our previous work \cite{bougaham2021ganodip}, ignoring the latent vector due to the decoder performance). The vector quantization mechanism is based on an embedding matrix of a discrete number of vectors to learn, resulting in a codebook $Z$. Its purpose is to provide vectors as close as possible to the images constituents, represented in the overall latent representations of the autoencoder. The encoded representation of the input image is therefore replaced by the nearest neighbor from the spatial collection of vectors learned, and is then decoded to a reconstructed image.

Finally, the reconstructed image is fed to the discriminator $D$ of a patchGAN \cite{isola2017image} (similar to a regular GAN but qualifying $N \times N$ patches, instead of the entire image through a single scalar). The discriminator objective is to collect a patch-wise reconstruction loss, giving information regarding its realness, for the training procedure. This way, the decoder part of the \textit{VQGAN} architecture takes the role of the generator part of the patchGAN, and the discriminator competes with it, stimulating the autoencoder and the codebook to provide realistic images, by receiving both the original and the reconstructed images. Its benefit is to provide images with high quality, instead of blurred ones that the \textit{VQ-VAE} suffer from \cite{esser2021taming}.\\

The end-to-end training procedure is guided by a combination of the reconstruction loss, the vector quantization loss, and the GAN loss. The reconstruction loss is a pixel-wise mean square error to capture detailed information, and a perceptual loss to capture semantic one: the original and reconstructed images are fed into a pre-trained VGG-16 network, and their last layer feature vectors MSE is computed. The global loss to optimize is defined as:
\begin{equation}
\mathcal{Q^*} = \underset{E,G,Z}{argmin} \, \underset{D}{max} \underset{{\text{x}} \thicksim p(x)}{\mathbb{E}}[\mathcal{L}_{VQ}(E,G,Z) + \lambda\mathcal{L}_{GAN}(\{E,G,Z\},D)],\label{eq:Loss}
\end{equation}
where
\begin{equation*} 
    \begin{cases}
        \begin{split}
        \resizebox{.85\linewidth}{!}{
          \begin{minipage}{\linewidth}
            \begin{align*}
                \mathcal{L}_{GAN}(\{ E,G,Z \},D) & = \Bigg[\log D(x) + \log \Bigg(1-D\bigg(G \Big(Z \big(E (x) \big) \Big) \bigg) \Bigg) \Bigg], \\\\
                \mathcal{L}_{VQ}(E,G,Z) & =  \bigg\| x- G \Big(Z \big(E (x) \big) \Big) \bigg\|^2 + \|sg[E(x)] - z_q\|^2_2 + \|sg[z_q] - E(x)\|^2_2, \\\\
                \lambda & = \frac{\nabla_{GL}[\mathcal{L}_{rec}]}{\nabla_{GL}[\mathcal{L}_{GAN}]+\delta},
            \end{align*}
          \end{minipage}}
        \end{split}
    \end{cases}
\end{equation*}
with $sg$ being the stop-gradient operation, $\mathcal{L}_{rec}$ being the perceptual loss capturing the differences between $x$ and \scalebox{0.7}{%
$ \text{G} \Big( \text{Z} \big( \text{E} \text{(x)} \big) \Big)$} with the pre-trained VGG-16 network, $\nabla_{GL}[.]$ the gradient of its inputs w.r.t the last L layer of the decoder and $\delta = 10^{-6}$ a small constant added for numerical stability \cite{esser2021taming}.

\subsubsection{GanoDIP Abnormal Candidates Isolation}\label{subsubsec2}
\begin{figure}[htbp]
    \centering
    \includegraphics[width=1\linewidth]{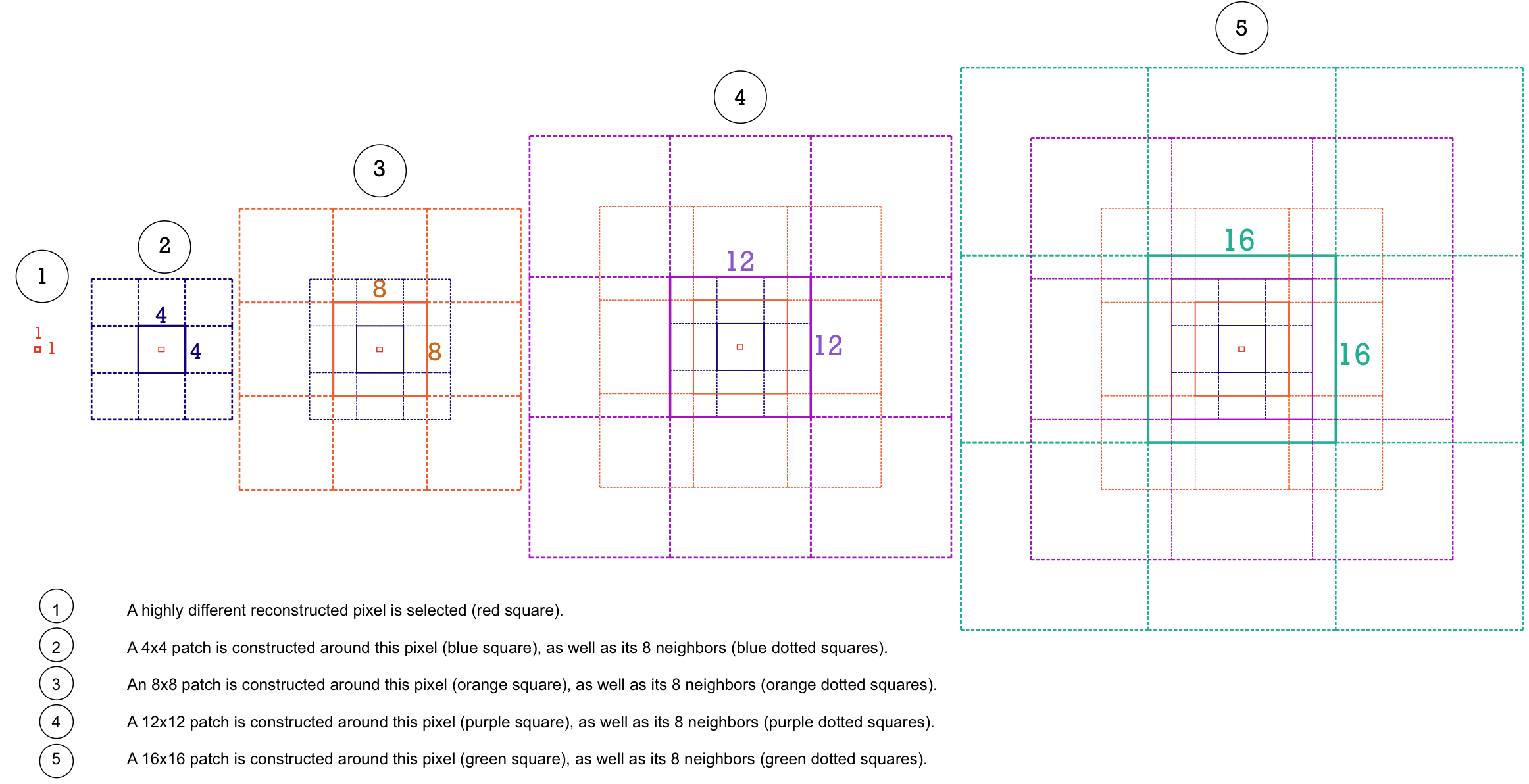}
    \caption{(Color online) Overview of the zoom-out-and-shift technique. $n \times 9$ patches are created (here $n=4$; enlarged factor $\alpha=4$) to focus on the $p$ most different pixels, at different scales.}
    \label{fig:ZoomShift}
\end{figure}

Once the reconstruction model is trained, we reconstruct the test set images. The \textit{GanoDIP} inference step we developed in our previous work \cite{bougaham2021ganodip} is applied by extracting the most different patches between the original and the reconstructed image. In this work, instead of considering only the highest MSE patch-wise, we first keep the $p$ pixels showing the highest absolute differences, then we construct patches by zooming out and shifting around these $p$ pixels. Figure~\ref{fig:ZoomShift} gives an overview of this technique.

The zoom-out-and-shift method is repeated $n$ times, enlarging the patches of $\alpha$ pixels each time, yielding different $s\times s$ patch sizes. The shift step allows constructing 8 more patches for each size in the neighborhood (top-left, top-center, top-right, center-left, center-right, bottom-left, bottom-center, bottom-right), covering an entire estimated abnormal set of pixels. According to the industrial partner experts, this method imitates the natural way of a human visual search, when an anomaly is expected in a small area. It assumes that, if a defect occurs, several close pixels are concerned instead of a single one. As reported in \cite{eckstein2011visual}, the normal areas are considered as a white noise, and the center–surround mechanism is applied to efficiently search and appreciate the defect.

These $n\times 9 \text{ (center + 8 neighbors)}$ patches repeated on the $p$ most different pixels are finally used to compute the Frechet Inception Distance \cite{heusel2017gans} (FID) between the original and the reconstructed images, considering their feature vector difference (generated with an Inception Resnet-V3 network, pre-trained with ImageNet) . This metric helps getting perceptual differences, pertinent with our computer vision task, instead of pure pixel differences, unrelated to the visual similarity between two images.

From these $p \times n \times 9$ FID values, we keep the $q$ highest as the estimated abnormal candidates. At the end, we get $q$ patches with different sizes, containing the highest perceptual difference between the original and the reconstructed image. They will first localize the estimated abnormal areas in the overall image and then be exploited for the second step of the methodology, the metrics collection described in the next subsection.

\subsection{Second step: Multi-level Difference Metrics Collection}\label{subsec4}
Recent anomaly detection methods (\cite{bougaham2021ganodip}, \cite{akccay2019skip}, \cite{schlegl2017unsupervised}, \cite{schlegl2019f}) design an anomaly score directly with regular pixel-wise or patch-wise MSEs, and the losses yielding from their neural network architecture. We aim to reproduce this strategy, in addition to taking into account the computer vision dimension of the task. Different type of distances between the input and the reconstruction will therefore be used to design the anomaly score, giving the method the best set of information on which to rely. Indeed, once the reconstruction has been performed and the most different patches have been identified on the test set, several multi-level metrics will be collected to characterize an anomaly present in the image. These metrics will be associated with each image, expressing a wide variety of information contained in the difference between the original and the reconstructed image.\\

Three different metric levels are considered. The first metric level describes the reconstruction quality image-wise. This is the case for the whole image pixel-wise MSE between the input and the reconstructed image (raw reconstruction loss), the pre-trained VGG-16 last layer MSE between the input and the reconstructed image features (raw perceptual loss), the MSE between the encoded latent representation and its quantized version (raw quantization loss), the patchGAN discriminator average loss for both the input and the reconstructed image (raw GAN losses), the ORB image matching difference \cite{karami2017image}, or the aggregated values of all the pixels resulting from the input-versus-reconstructed absolute difference. These aggregated values are the sum, maximum, minimum, mean, first quartile, second quartile (a.k.a. the median), and third quartile.

The second metric level describes the reconstruction quality pixel-wise. In this case, the goal is to retrieve information from the $p$ highest pixel values resulting from the input-versus-reconstructed absolute difference. The same aggregated values are considered to qualify this set of most different pixels.

\begin{figure}[htbp]
    \centering
    \includegraphics[width=0.9\linewidth]{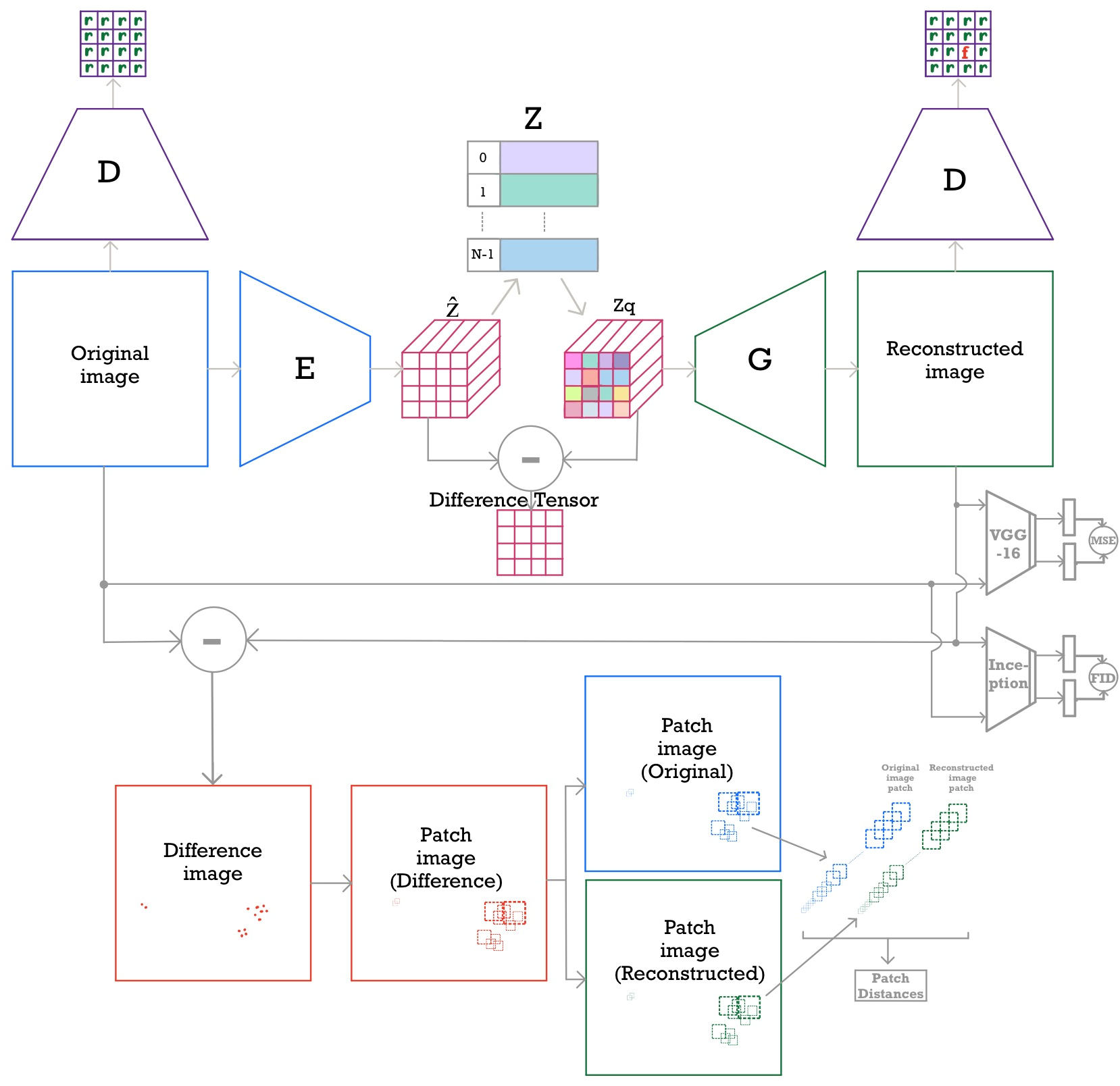}
    \caption{(Color online) Overview of the \textit{VQGanoDIP} architecture, at the inference step. A multi-level metric collection is extracted from the input and reconstructed images, containing insightful information to determine whether an anomaly is present or not.}
    \label{fig:VQGAN_inference}
\end{figure}

The third and last metric level describes the reconstruction quality patch-wise. The methodology involves computing matrix distances on the $q$ patches selected via the $p$ highest pixel values and the zoom-out-and-shift technique. We append each of these patch distance values into a sequence and apply the aggregated method to get a scalar that qualifies this set of patches. This is the case for several established distances between two matrices (Frechet, SSIM, Braycurtis, Canberra, Euclidian, Cosine, Wasserstein, Hamming, Minkowski, Jensen–Shannon divergence, etc.), as well as for the FID (triggering the selection). Figure~\ref{fig:VQGAN_inference} shows the architecture at the metric collection step.

For a dataset with the same type of images and a fixed position (like the PCBA one, where the same product model does not vary in rotation or translation), the same set of metrics, weighted by the normal variation inside the majority class, is computed. This technique brings business knowledge to the method, making it possible to reduce the distances where many normal variations are already observed in a normal class dataset. To do so, we isolate $m$ unseen normal images, compute and normalize the pixel-wise average difference value between the input and the reconstructed images. We then obtain a mask (after a flip operation) reflecting the average difficulty that the reconstruction model encounters, due to the high normal variations. If a pixel varies a lot in the normal images, the mask pixel value will be close to zero. Otherwise, it will be close to one. These mask pixel values will be multiplied by the difference pixel values between a test input and reconstructed image, and will weight the difference computed (especially on high normal variation areas) with the help of a part of the business knowledge. Then the FID zoom-out-and-shift and the metrics collection are performed, based on these new $p$ most different pixels. That doubles the metrics number (with and without the weighting mask), which will be used during the last step, detailed in the next subsection.

\subsection{Third step: Composite Anomaly Score Creation}\label{subsec5}
Associating an anomaly score to each test set image is the last step of the methodology. At this stage, we will use the few abnormal images we have at hand to create a classifier able to discriminate between the two classes. Indeed, instead of only using the reconstruction and eventually the latent loss as it is usually performed in anomaly detection techniques, we will feed a dataset built upon the metrics collected (instances in rows, metric values in columns) into a classifier, in order to let it build a new, composite, anomaly score that best discriminates the classes. \\

The anomaly score is designed in 4 phases. A data processing first phase is applied, removing features with a constant value. If several metrics are highly correlated (more than 95\%), only one of them is kept. Values are also scaled into a [0\,-\,1] range. 

Then, in a second phase, a randomized search for the hyperparameters optimization, efficient when the number of hyperparameters is large \cite{bergstra2012random}, is applied to the entire dataset to achieve the best accuracy. If the dataset is imbalanced (this is the case for the MVTEC-AD datasets), we rely on the SMOTE algorithm \cite{chawla2002smote} to generate artificial positive samples (minority class) in the cross-validation stage, keeping pertinent the accuracy metric to optimize. 

The third phase is a Leave-One-Out Cross-Validation (LOOCV) procedure, with the best-resulted hyperparameter set. This technique is computationally heavy, but even if we have a dataset composed of twice the number of the minority-class images (we take the number of normal images as the same we have for the abnormal ones), it is worth training as many models as there are instances, to get a reliable assessment. A stratified k-fold cross-validation is performed on the classifier, splitting all the instances (except one) in a training and validation set. A grid search procedure selects the best classifier type between several binary classification ones, namely a decision tree (DT), a random forest (RF), an extra trees classifier (ET), an adaptive boosting classifier (ADA), a light gradient boosting machine (LGBM), a gradient boosting classifier (GBC), an extreme gradient boosting classifier (XGBoost), a logistic regression (LR), a K nearest neighbor classifier (KNN), a gaussian naive bayes classifier (NB), a linear discriminant analysis (LDA) and a quadratic discriminant analysis (QDA). 

In the final phase, we set the threshold as the lowest prediction probability that the classifiers associate to the abnormal test set. This way, we ensure that all the abnormal images in the test set are predicted as it has to be, and we can evaluate the classifier through the accuracy, being degraded only with the normal images misclassification. Therefore, the prediction probability that the classifier gives to a test image to be abnormal is the composite anomaly score.

\section{Evaluation}\label{Evaluation}

To summarize the above section, the proposed methodology is composed of three steps, which are a reconstruction model creation and anomalies localization, a metrics collection based on the input and reconstructed images, and a binary classifier training with the objective of building a composite anomaly score. After having detailed the methodology, this section is devoted to the experimental setup followed by the qualitative and quantitative model evaluation, finally discussed.

\subsection{Experimental Setup}\label{subsec6}

For implementation purposes, several hyperparameters have to be tuned. In practice, we observed that the following decisions are the best to deal with the model performance and inference time. These following values are tailored for the PCBA dataset, but also tested on the  MVTec-AD datasets \cite{bergmann2019mvtec}.

In order to capture a large amount of insightful information on the PCBA dataset, we increased the default number of the codebook entries from 1024 to 2048 and its dimensionality from 256 to 512. This way, a large variety of textures, reflections, orientations and shapes can be captured by the codebook, and returned to the quantized latent representation. 

For the PCBA dataset, 360 $1024 \times 1024$ anomaly-free images are randomly selected to create the first step reconstruction model. The imbalanced nature of the dataset constrains us to work with only a few abnormal class images, compared to many normal class ones at our disposal. We have 174 abnormal images, so we randomly select 174 normal images (easily available, as it is the majority class) to get a balanced dataset of 348 images for the last-step composite anomaly score model creation. For the  MVTec-AD datasets, half of the normal images have been selected for the first step reconstruction model training. The other half of the normal images and all the abnormal images were selected for training and evaluation of the last step anomaly score model.

On these test images, we apply the \textit{GanoDIP} inference step as we developed in our previous work \cite{bougaham2021ganodip}, keeping the $p=100$ highest absolute difference pixels. The zoom-out-and-shift step is repeated $n=4$ times, with an enlarged area of $\alpha=4$ pixels each time. We therefore get 36 patches of $4\times4$, $8\times8$, $12\times12$ and $16\times16$ sizes. We finally keep the $q=250$ worst FID patches, between the original and the reconstructed images, as the estimated abnormal candidates. 

To evaluate areas of high normal variability, the weighting mask is based on $m=30$ unseen random normal images and is only suitable for the PCBA dataset. Indeed, in the MVTec-AD datasets, objects and textures are rotated or translated, preventing the possibility of considering such a weighting mask.  

The cross-validation to train the anomaly score classifier is a 5-fold stratified one. The randomized search for the optimal hyperparameters is executed for 500 iterations. The classifier selected is the one that gives the best accuracy in these conditions. 

All experiments have been undertaken with an Nvidia RTX A6000 GPU, an Intel i7 CPU, using Python 3.8, Cuda 11.2 and Pytorch 1.10.

\subsection{Qualitative Assessment }\label{subsec8}

The proposed methodology is qualitatively assessed, through the reconstruction quality (Sections~\ref{subsubsecQualiPCBA} for the private real-world PCBA dataset and~\ref{subsubsecQualiMVTec} for the public MVTEC-AD datasets), and through a Visual Turing Test for the PCBA in Section~\ref{subsubsecVTTPCBA}.

\subsubsection{PCBA Reconstruction Quality}\label{subsubsecQualiPCBA}

\begin{figure}[h]
    \centering
    \includegraphics[width=1\linewidth]{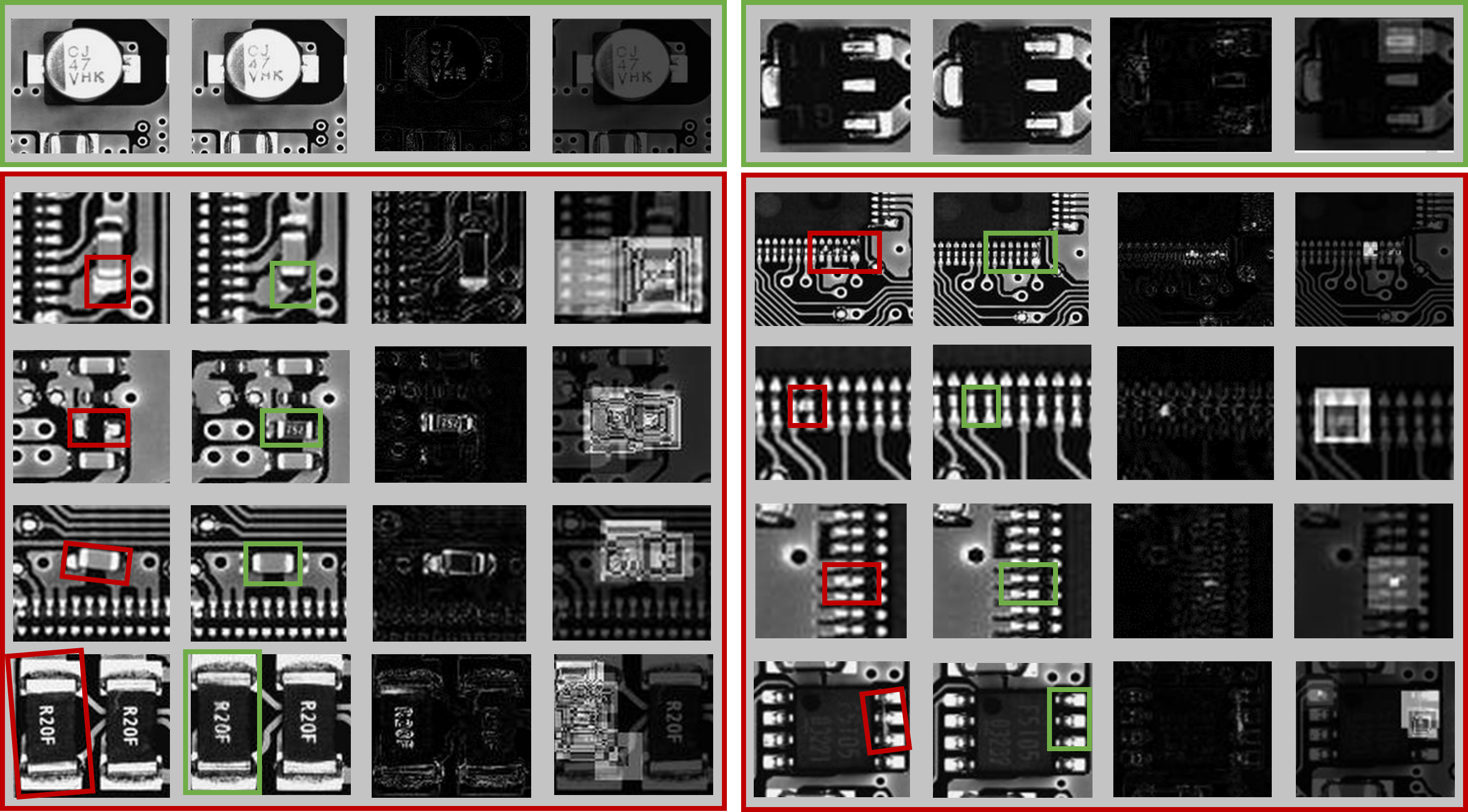}
    \caption{(Color online) Two sets of original images (1st column) with the defect highlighted (red frame), reconstructed images (2nd column) with the recovered pixels highlighted (green frame), difference images (3rd column) and patch images (4th column) for 10 different zoomed areas ($\approx X10$) of the PCBA dataset (placed in rows). The first row shows anomaly-free areas unlike the four last rows, where a defect is observed. The left set shows component defects, and the right set shows solder defects.}
    \label{fig:Quali1}
\end{figure}

A first way to evaluate the anomaly detection method proposed is to visually check how the original defects are recovered. Figure~\ref{fig:Quali1} shows several zoomed area examples where PCBA images contain anomalies that are correctly recovered, thanks to the \textit{VQGAN} implementation.
The reconstruction, difference, and abnormal patches estimation images are presented. The left set of images illustrates very small component shifts or absence, and the right set shows slight solder defects. We can see that the defects are correctly recovered (2nd column of each sets), proving the reconstruction efficiency for such complex data. The input data distribution is well followed for normal areas (first row of the figure), yielding low pixel differences (3rd column of each sets) after reconstruction. This statement means that the reconstruction model does not fall into the posterior collapse problem as it was the case in our previous work \cite{bougaham2021ganodip}. The probable cause was a stronger generator that always generated a quasi-identical image for any input image, whatever the latent representation variations. This limitation, which yielded a unique golden sample, is now solved.

Another step forward is the possibility of reconstructing very small anomalies, thanks to the $1024 \times 1024$ resolution. The \textit{VQGAN} architecture makes it possible to consider an entire high-resolution input image at once, instead of patching it with a lower resolution, as it is the case for the \textit{f-AnoGAN} \cite{schlegl2019f} method, for instance. Therefore, difficulties of a challenging dataset like the PCBA one (small components in an information-rich global image) in the industrial context (entire image needed to evaluate the method with confidence before going into production) can be overcome.

\begin{figure}[h]
    \centering
    \includegraphics[width=0.85\linewidth]{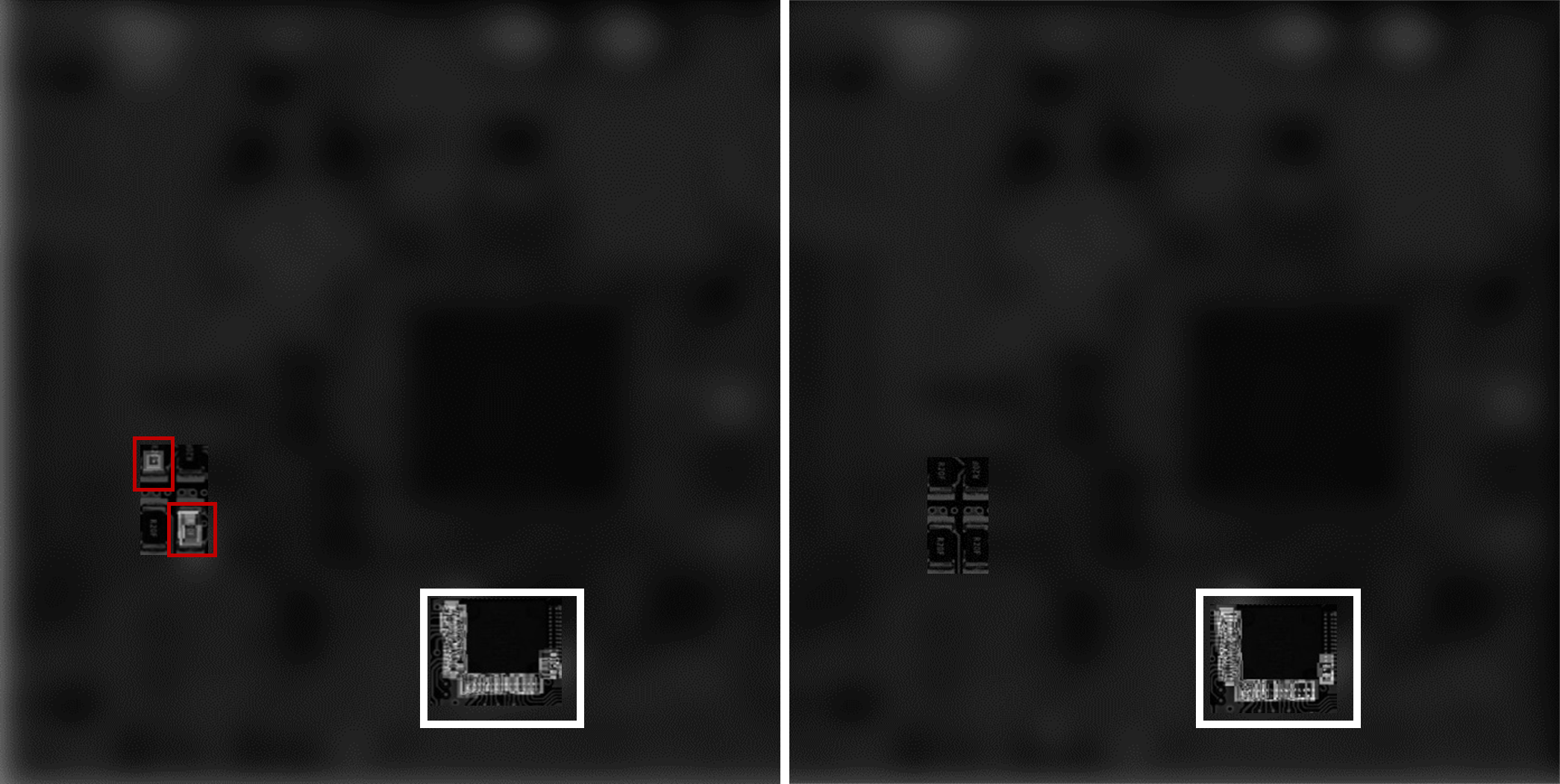}
    \caption{(Color online) Abnormal patches estimation with (left image) and without (right image) the weighting technique. The abnormal patches estimation (false positives on the left unblurred red-framed area) disappears with this technique, focusing on true positives (right unblurred white framed area). Some parts of the images have been anonymized (material under intellectual property).}
    \label{fig:Quali2}
\end{figure}

Regarding the overall difference images, we can state that some specific zones are always noisy due to the normal variability of the dataset (marking on the components, reflections on large solder pads, details on the 2D serial number barcode that highly change, image by image). However, thanks to the weighting method, these false-positive differences are reduced, letting the \textit{GanoDIP}-like technique choose the true positives. Figure~\ref{fig:Quali2} shows the relevance of this technique, which can only be applied with an adjusted position pre-processing task (here thanks to fiducial reference centering).
We can appreciate how the focus on abnormal areas is improved when the weighting technique is applied, selecting the suitable $p$ most different pixels (and thus the right $q$ highest FID patches), by reducing this difference when a high normal variation has been previously observed. This improvement is important when distance metrics will be used for the composite anomaly score (signal noise ratio increased).\\

\begin{figure}[h]
    \centering
    \includegraphics[width=1\linewidth]{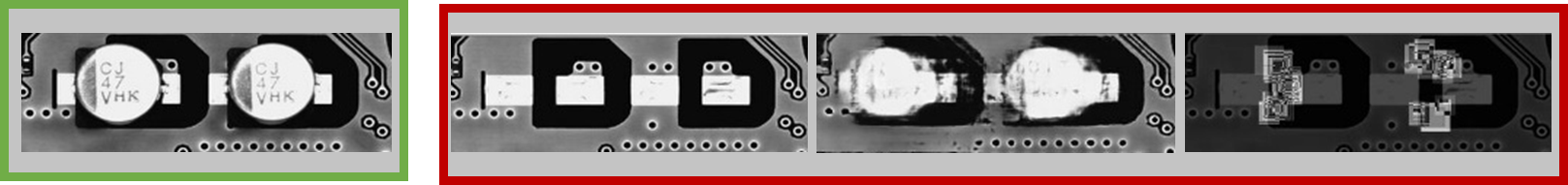}
    \caption{(Color online) Zoomed area ($\approx X10$) of a normal product image (left-green-framed 1st image) and an abnormal product image (right-red-framed image). One can see in the abnormal image that the input (2nd image) presents a pretty large defect (absence components), reconstructed with artefacts (3rd image) but still well patched (4th image).}
    \label{fig:Quali3}
\end{figure}

Despite these promising observations, some limitations remain. Figure~\ref{fig:Quali3} illustrates that for larger anomalies, the reconstruction is not as efficient, showing artifacts in the abnormal set of pixels. 
We can reasonably think that this comes from the latent representation dimensions, well fitted to deal with smaller details. Indeed, the choice for the codebook entries of 2048 and its dimensionality of 512 is particularly pertinent to catch the very small details in the image (like solder bridges or little component shifts) and the fine texture rendering (the PCBA silk or the solder pads reflects). This leads to difficulties for larger defect reconstructions. However, the artifacts reconstructed present all the same significant differences compared to the original abnormal image, and the \textit{GanoDIP}-like patch isolation method can still focus on the right area.\\

The quality reconstruction is a key indicator, but it is not sufficient to estimate the methodology performance. Another indicator is the location correctness of the abnormal estimated patches. The difficulty lies in the fact that all the $q$ patches will compete with each other to reveal the anomaly, and it is useful to notify that the first condition is to get at least one patch on the defect. For this concern, after the test set reconstruction, we observe that all the defects of the PCBA dataset are covered by a patch, which will be used as insightful information for the anomaly score creation.

\subsubsection{PCBA Visual Turing Test}\label{subsubsecVTTPCBA}

Another way to assess the methodology qualitatively is to make a Visual Turing Test (VTT) on a normal set of PCBA images. Indeed, to implement the method in a real-world production line, domain experts need a high degree of confidence. The \textit{VQGAN} model is hardly explainable (due to the multiple deep neural networks), but the VTT could reassure the experts on the reconstruction quality, being the first important block of the entire methodology. Therefore, this test has been conducted with 5 experts with different expertise levels, used to manipulate the PCBA images. 

The protocol followed has been inspired by classical ones, reported in \cite{salimans2016improved}, \cite{han2018gan} or \cite{schlegl2019f}. 50 original normal images and 50 other reconstructed ones are randomly shuffled and shown to the experts. 16 seconds are given to the participants to determine if the image shown is real (original from the production line camera) or fake (reconstructed by the model). Between 2 images, 5 seconds of a blank screen is displayed to reset the visual memory and encode the judgment in a document. In the middle of the test (50th image), a 5 minutes break is taken to keep them focused until the end. No discussion between them is possible to avoid any eventual bias. 

If the candidates cannot clearly state whether the images are artificial or not, then we can conclude that the reconstruction model generates realistic images, thanks to the architecture performance. This would be a good sign that this first reconstruction block is satisfying and brings confidence in the overall methodology. This procedure is a kind of human (expert) discriminator, like the one we have in the GAN part of the reconstruction model.\\ 

\begin{figure}[h]
    \centering
    \includegraphics[width=0.9\linewidth]{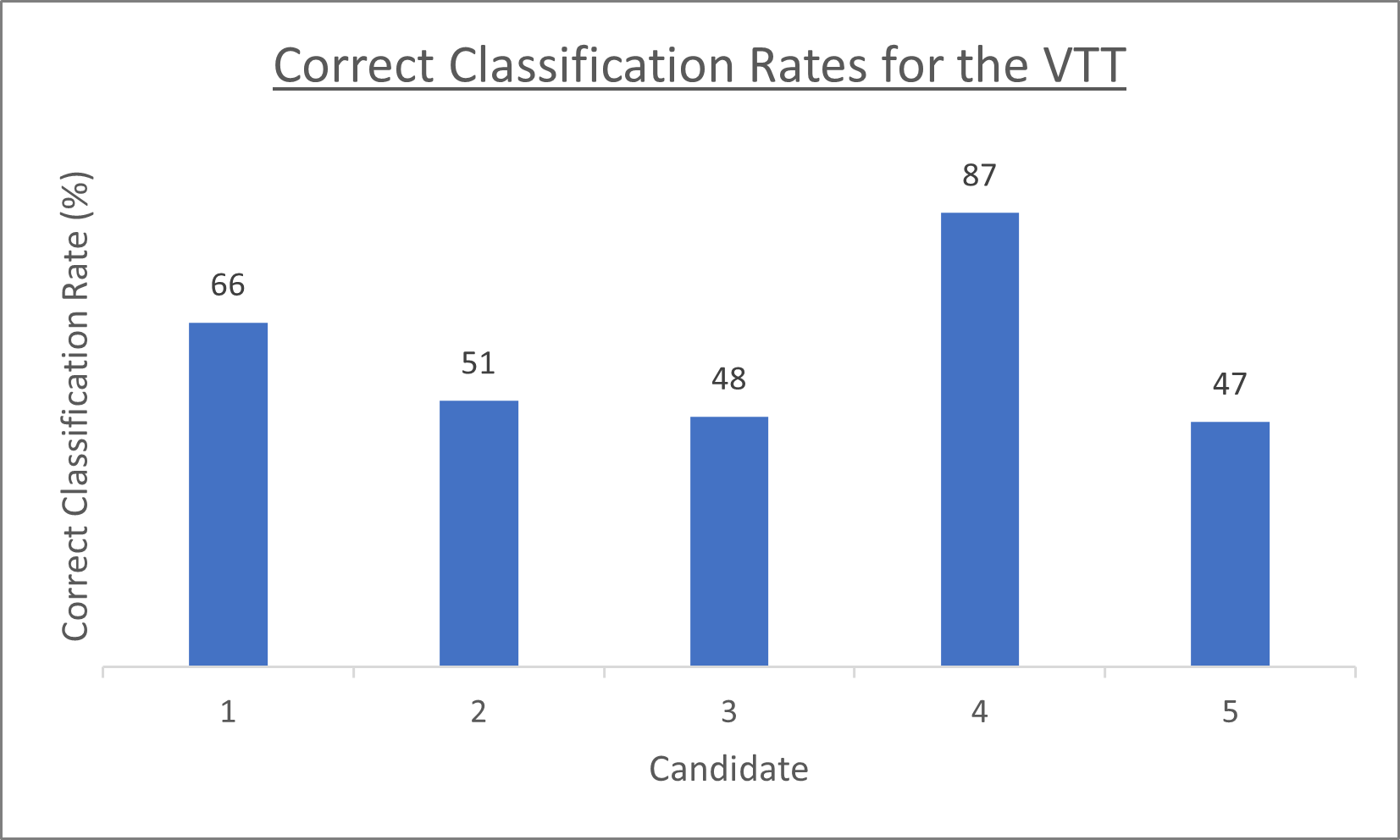}
    \caption{Classification Rates for the 5 domain experts participating in the VTT.}
    \label{fig:VTT_Result}
\end{figure}

The results are presented in the Figure~\ref{fig:VTT_Result}.
The correct classification average rate of 59.8\%, with a 13.9\% standard deviation, proves the difficulty of a domain expert distinguishing real images from fake ones. We can therefore conclude that the model generates high-fidelity images, a key argument for the users to adopt the algorithm.

\begin{figure}[h]
    \centering
    \includegraphics[width=0.7\linewidth]{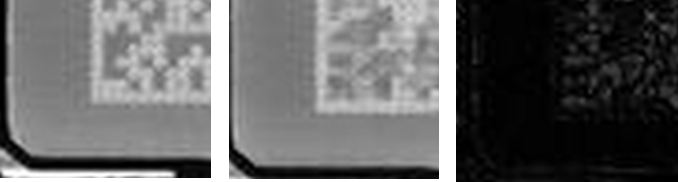}
    \caption{Zoomed area ($\approx X20$) of the 2D serial number barcode where the original image (left image) gives a blurred reconstruction (middle image) with several pixel differences (right image).}
    \label{fig:VTT}
\end{figure}

An outlier stands out from the VTT results, with the highest score of 87\%. The candidate with this score is a computer vision specialist responsible for designing anomaly detection algorithms for another production process. We can therefore understand why his biased attention differs from other participants. After a debrief session with him, it appears that some areas were insightful in judging the realness of the images. This is particularly the case for the 2D serial number barcode area, as shown in Figure~\ref{fig:VTT}. 
Indeed, there are so many details and normal variations in this area that the model cannot reconstruct the dots composing the barcode matrix clearly. This yields to pixels that are smoothed, with a blurred effect. Hopefully, this limitation is not impactful for the anomaly detection because the weighting technique will reduce the difference pixel values. In addition, the defect opportunity in this area is very low.

\subsubsection{MVTEC-AD Reconstruction Quality}\label{subsubsecQualiMVTec}

The quality reconstruction can also be assessed on the popular MVTEC-AD datasets to figure out the genericity of the \textit{VQGanoDIP} methodology. This work focuses on the $1024 \times 1024$ images resolution. Therefore lower resolution datasets were discarded. 

\begin{figure}[h]
    \centering
    \includegraphics[width=0.9\linewidth]{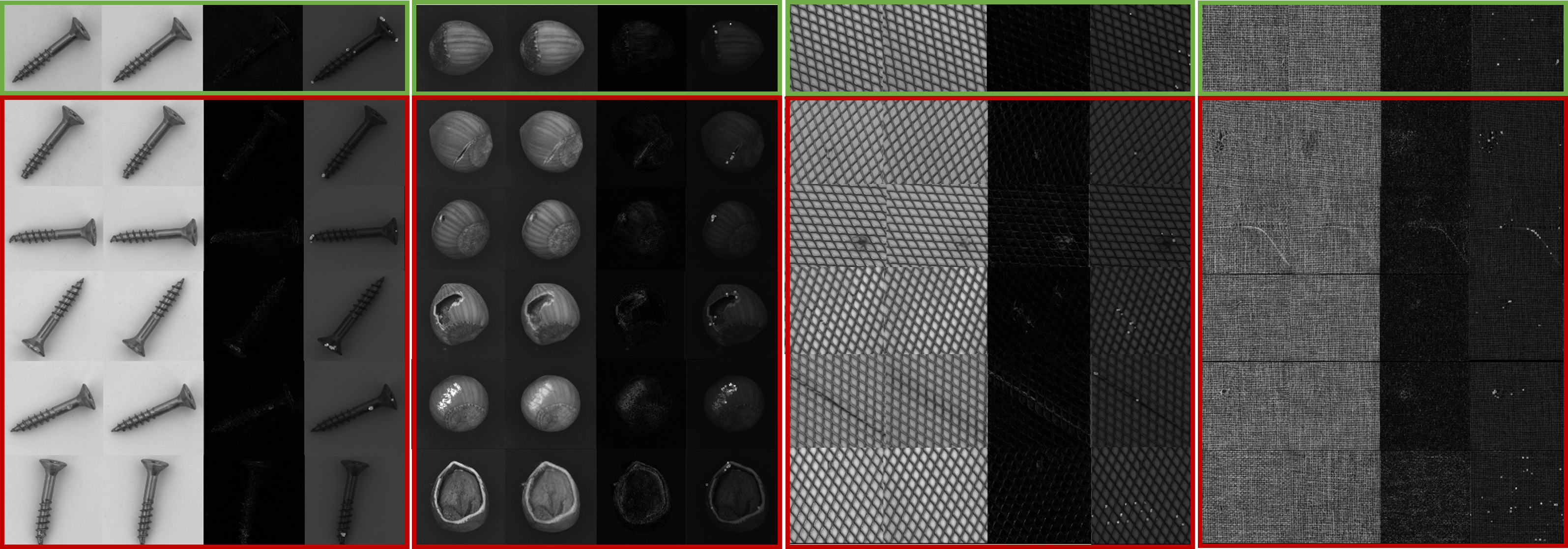}
    \caption{(Color online) One normal image (1st row) and four abnormal images (4 last rows) of the screw (1st block), the hazelnut (2nd block), the grid (3rd block) and the carpet (4th block) datasets. For each block, the original (1st column), the reconstruction (2nd column), the difference (3rd column) and the patch (4th column) images are shown.}
    \label{fig:Quali4}
\end{figure}

Figure~\ref{fig:Quali4} shows the original, reconstruction, difference, and patch images for some examples of object and texture products, namely the screw, the hazelnut, the grid, and the carpet datasets. 
We can confirm that the small defects are better recovered than the larger ones from this figure. For instance, a screw (1st block images) with a broken tail (2nd and 3rd rows) is more difficult to reconstruct than a scratch on the head (4th row) or the neck (5th and 6th rows). This is also the case for the hazelnut (2nd block images), where a rough crack (4th and 6th rows) is not well recovered, unlike a small cut (2nd row) or hole (3rd row). Despite these reconstruction difficulties, the abnormal estimated patches can still focus on the abnormal areas, even if the clustering effect is less observable. 

Concerning the texture images, the same observation can be done (difficult reconstruction for rough defects but still estimated as abnormal), in addition to a larger split of the patches in the overall image. This is due to the high normal variations that offer the texture images (orientation, fibers, etc.), competing with the true-positive areas. We can therefore conclude that object images have better reconstruction performance than texture ones. This is due to the predominance object dimension of the PCBA images (compared to the texture dimension), which required hyperparameters selection adapted to this feature.

\subsection{Quantitative Assessment}\label{subsec7}

A second way to assess the anomaly detection methodology is to measure established classification metrics, namely the accuracy, the precision, and the false-positive rate. In the case of imbalanced datasets, precision is a more relevant metric because it does not include true negatives (the majority class). There is no interest in measuring other metrics like the sensitivity (also named recall or true-positive rate), which will always be equal to 1 due to the absence of false negatives under the zero-false-negative constraint, or the AUCROC, as the only interesting threshold for our business case is the one able to detect all abnormal instances. As this specific anomaly score threshold cancels the false negatives, the confusion matrix is asymmetric, with the entire misclassified instances being false positives, giving the accuracy directly. The classification metrics are presented and discussed for the PCBA and the MVTEC-AD datasets.

\subsubsection{PCBA Classification Metrics}\label{subsubsecQuantiPCBA}

Table~\ref{PCBA_metric_table} summarizes the metrics under the zero-false-negative constraint (ZFN columns), and, in a standard way, without this constraint (STD columns), on the PCBA dataset.
\begin{table}[h]
\begin{center}
\begin{minipage}{\textwidth}
\caption{\textit{VQGanoDIP} classification metrics for several classifiers on the
PCBA dataset, considering the zero-false-negative constraint (ZFN columns) or not (STD). The $\uparrow$ sign means the highest the best, unlike the $\downarrow$ sign means the lowest the best. Values are sorted with the ZFN Accuracy column, and bold ones are the best of each column.}\label{PCBA_metric_table}
{\setlength{\tabcolsep}{0pt}
\begin{tabular*}{\textwidth}{@{\extracolsep{\fill}}lccccccccc@{\extracolsep{\fill}}}
\toprule%
& \multicolumn{2}{@{}c@{}}{Accuracy(\%)$\uparrow$} & \multicolumn{2}{@{}c@{}}{Precision(\%)$\uparrow$} &  \multicolumn{2}{@{}c@{}}{FPR(\%)$\downarrow$} & \multicolumn{2}{@{}c@{}}{FNR(\%)$\downarrow$} \\\cmidrule{2-3}\cmidrule{4-5}\cmidrule{6-7}\cmidrule{8-9}%
Classifier & STD & ZFN & STD & ZFN & STD & ZFN & STD & ZFN
\\\cmidrule{2-2}\cmidrule{3-3}\cmidrule{4-4}\cmidrule{5-5}\cmidrule{6-6}\cmidrule{7-7}\cmidrule{8-8}\cmidrule{9-9}%
ET &
\textbf{95.69} & \textbf{87.93} &
\textbf{98.18} & \textbf{80.56} & 
\textbf{1.72} & \textbf{24.14} & 
6.9 & 0 \\
XGBoost &
91.67 & 83.05 & 
89.19 & 74.68 & 
11.49 & 33.91 &
\textbf{5.17} & 0\\
RF &
94.25 & 81.03 & 
97.53 & 72.5 & 
2.3 & 37.93 & 
9.2 & 0 \\
GBC &
93.68 & 79.31 & 
96.91 & 70.73 & 
2.87 & 41.38 & 
9.77 & 0 \\
LGBM &
93.1 & 75.57 & 
95.18 & 67.18 & 
4.6 & 48.85 & 
9.2 & 0 \\
ADA &
93.39 & 73.28 & 
95.21 & 65.17 & 
4.6 & 53.45 &
8.62 & 0 \\
LR &
93.97 & 53.45 & 
94.74 & 51.79 & 
5.17 & 93.1 & 
6.9 & 0 \\
QDA &
90.52 & 50 & 
96.08 & 50 & 
3.45 & 100 & 
15.52 & 0 \\
KNN &
91.38 & 50 & 
97.37 & 50 & 
2.3 & 100 & 
14.94 & 0 \\
LDA &
94.83 & 50 & 
95.88 & 50 & 
4.02 & 100 & 
6.32 & 0 \\
DT &
89.94 & 50 & 
93.17 & 50 & 
6.32 & 100 & 
13.79 & 0 \\
NB &
81.9 & 50 & 
94.4 & 50 & 
4.02 & 100 & 
32.18 & 0 \\
\botrule
\end{tabular*}}
\end{minipage}
\end{center}
\end{table}
We can conclude that the Extra Tree Classifier offers the best classification metrics under the ZFN constraint, with an accuracy of 87.93\% and 95.69\% without this constraint (with a larger false-negative rate than a false-positive rate in this case). Notice that the the false-negative rate metric under the ZFN column is zero for all the classifiers, due to our quality exigence constraint. The linear and quadratic discriminant analysis (LDA, QDA), K nearest neighbor (KNN), decision tree (DT) and gaussian naive bayes (NB) classifiers raise 50\% (the dataset imbalance rate value) of Accuracy and Precision, and 100\% of false-positive rate, under the ZFN constraint. This situation happens because they encounter at least one challenging image for which the training failed to capture the discriminating features. Therefore, the lowest prediction probability for the positive class is zero, meaning that the classifier judges at least one positive instance with a 0\% confidence to be positive. It yields an anomaly score threshold of zero, and all the instances will be classified as positives. We can conclude that these classifiers are not powerful enough to be used as our dataset anomaly score. The ensemble decision trees family seem much well fitted to the task.

\begin{figure}[h]
    \centering
    \includegraphics[width=0.6\linewidth]{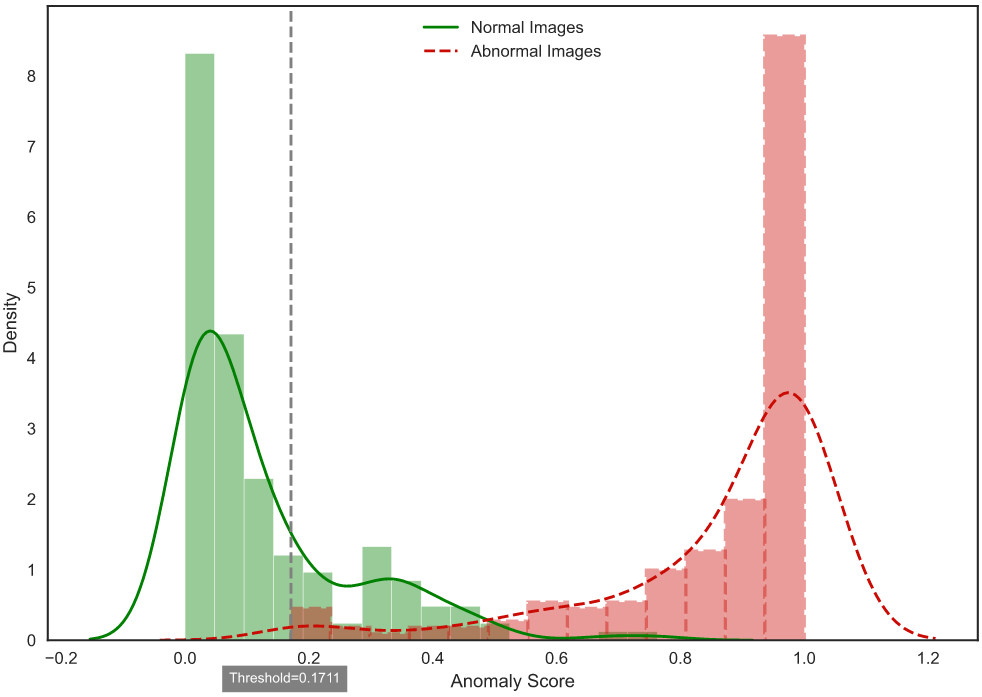}
    \caption{(Color online) Anomaly score distributions of normal (solid-green line and bars) and abnormal (dashed-red line and bars) images for the PCBA dataset, with the threshold value (vertical dashed line in grey) that satisfies the ZFN constraint.}
    \label{fig:AS_Score}
\end{figure}

Figure~\ref{fig:AS_Score} shows the anomaly score distributions for the normal and the abnormal images of the PCBA dataset, built with the best classifier, raising a 87.93\% accuracy. 
We can see the distributions are well separated (low anomaly score for the normal images and high anomaly score for the abnormal ones), even if the threshold (ensuring no missed detection) prevents an optimal split between them. It generates an overlap of the negative distribution, being the false positives (normal images scored above the threshold).

\begin{figure}[h]
    \centering
    \includegraphics[width=0.9\linewidth]{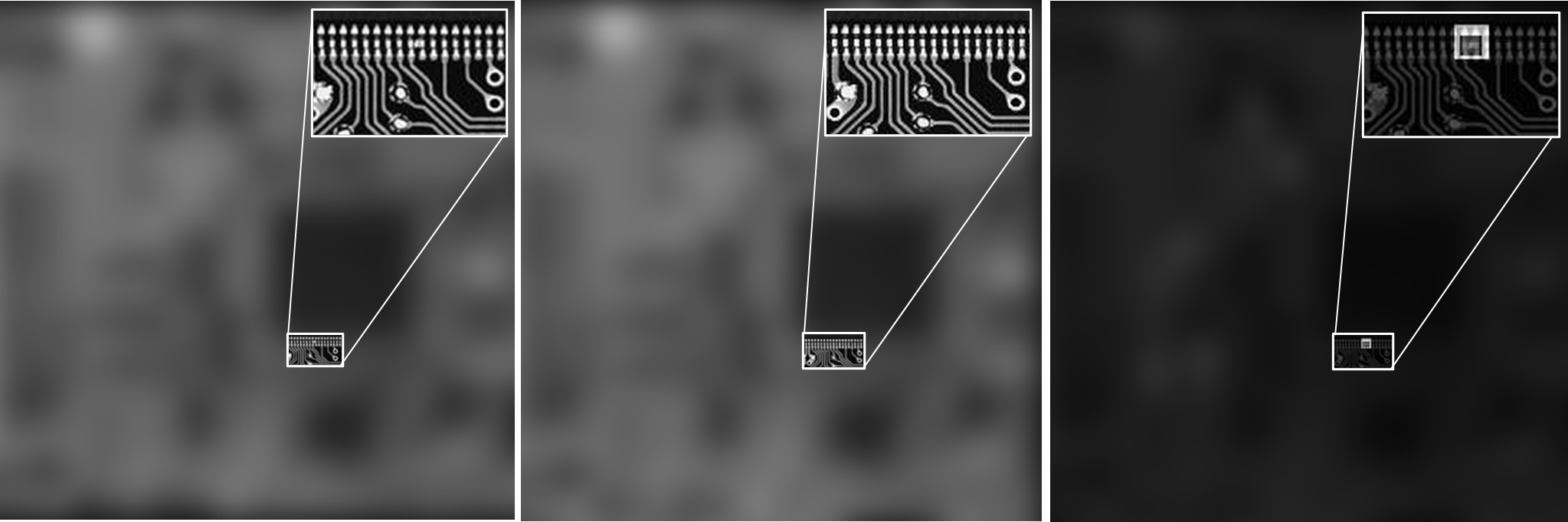}
    \caption{Original (1st column), reconstructed (2nd column), and patch image (3rd column) with a zoomed view ($\approx X10$) of the PCBA most challenging image. Some parts of the images have been anonymized (material under intellectual property)}
    \label{fig:Low_Img}
\end{figure}

The zoomed abnormal area of the image conditioning the adjusted threshold (anomaly score of 0.1711) is presented in Figure~\ref{fig:Low_Img}. 
This figure shows how difficult it is for the method to associate a high anomaly score with this very small defect, with respect to a low score for small normal variations. Indeed the small solder defect (a few $\text{mm}^2$ surface) in the input image is correctly recovered and an abnormal patch highlights the area thanks to the method. Nevertheless, this area competes with many other normal areas, giving difficulties for the classifier to build its decision function.

\begin{figure}[h]
    \centering
    \includegraphics[width=0.8\linewidth]{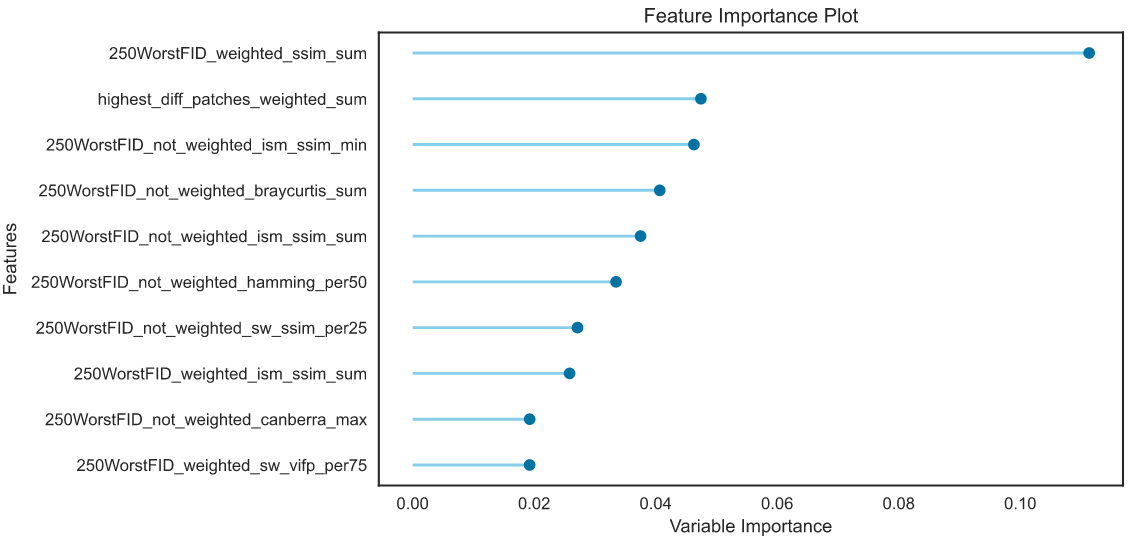}
    \caption{10 most infuent collected metrics that impact the discrimination decision.}
    \label{fig:Feat_Imp}
\end{figure}

It is also interesting to determine the most influent collected metrics that impact the discrimination decision. Figure~\ref{fig:Feat_Imp} shows the importance of the \textit{GanoDIP}-like zoom-out-and-shift FID patches (for the 10 most influent metrics, 9 of them rely on these patches), as well as the weighting technique ($\frac{4}{10}$ most influent metrics) and the SSIM metric ($\frac{5}{10}$ most influent metrics). 
We can also notice that the most efficient aggregation technique is the sum of all the patch distances  ($\frac{5}{10}$ most influent metrics). Finally, we can see that the most influent metric is the sum of the Structural SIMilarity (SSIM) values of all abnormal patches estimated, after applying the weighting mask to reduce the normal variations influence. It demonstrates the importance of considering the computer vision dimension in this task.\\

From a business point of view, each misclassification generated by the algorithm requires an operator visual inspection. Also, human misjudgement risks are proportional to the quantity to inspect. Therefore, an improvement on the accuracy directly impacts the time waste and the quality risk for the industrial partner. Here, thanks to the accuracy reached by the \textit{VQGanoDIP} approach, the average inspection time required is decreased from 8 seconds to 1.9 second, and the operator misjudgment rate is divided by a factor of 4. This represents a significant improvement and is definitely promising to keep the partner competitive.

\subsubsection{MVTEC-AD Classification Metrics}\label{subsubsecQuantiMVTec}

\begin{figure}[h]
    \centering
    \includegraphics[width=0.95\linewidth]{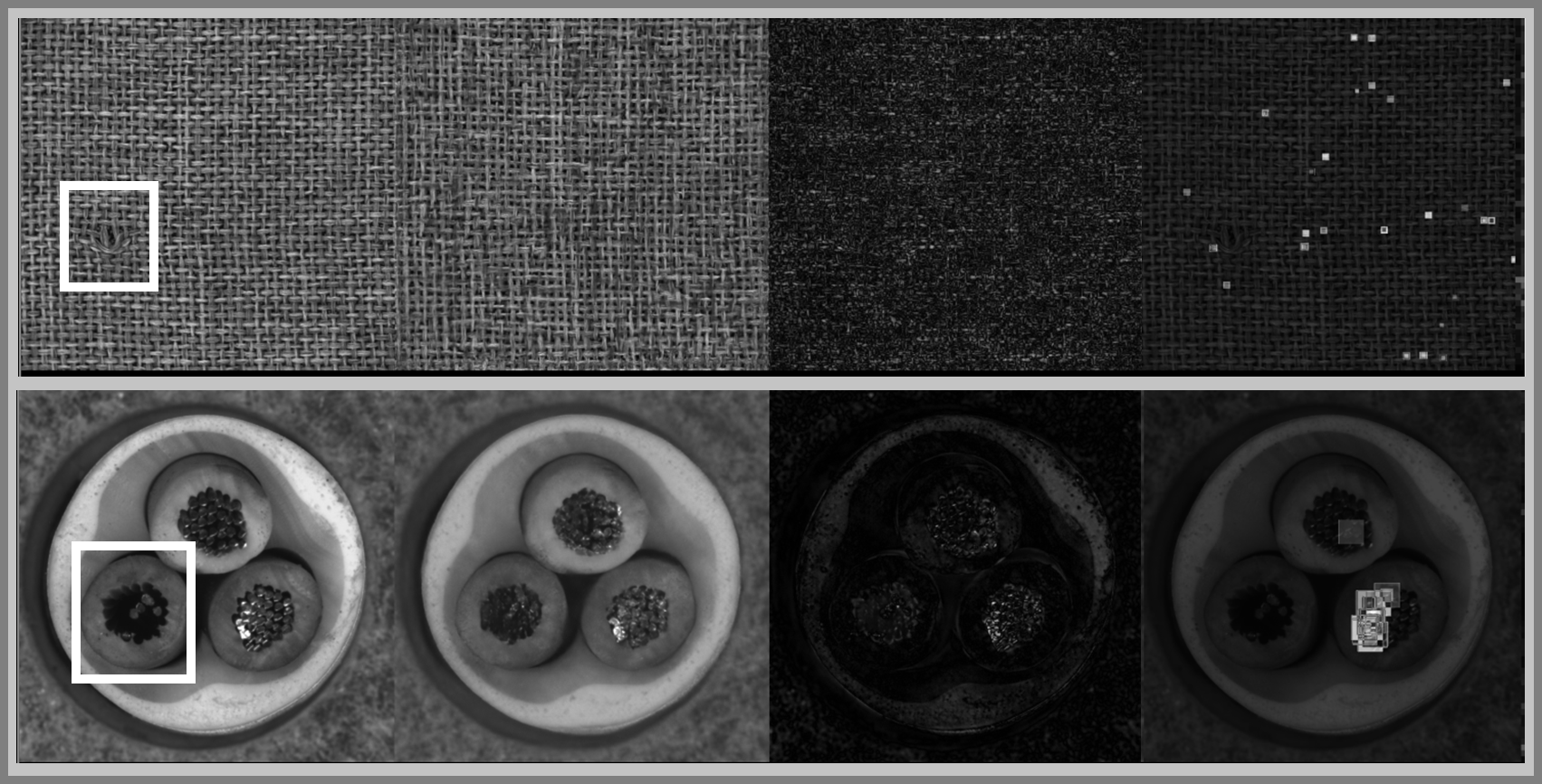}
    \caption{Original (1st column) with the anomaly white-framed, reconstruction (2nd column), difference (3rd column), and patch (4th column) images for Carpet (1st row) and Cable (2nd row) image example. For the Carpet, we can see, on the difference image, all the small normal variations that the model could not well reconstruct. This yields, in the patch image, in many small patches everywhere but not in the anomaly area. For the Cable, we can see that the missing wires cannot be well recovered, which fools the patches focus.}
    \label{fig:Quanti1}
\end{figure}

The same study has been performed on the public MVTec-AD datasets, especially on the images at the $1024 \times 1024$ resolution. 
The final summary of all the datasets, keeping the most accurate classifier under the ZFN constraint, is presented in Table~\ref{All_metric_table}.
\begin{table}[h]
\begin{center}
\begin{minipage}{\textwidth}
\caption{\textit{VQGanoDIP} classification metrics for the most accurate classifiers (indicated into parenthesis) on the PCBA and the $1024 \times 1024$ MVTec-AD datasets, considering the zero-false-negative constraint (ZFN columns) or not (STD). The $\uparrow$ sign means the highest the best, unlike the $\downarrow$ sign means the lowest the best.}\label{All_metric_table}
{\setlength{\tabcolsep}{0pt}
\begin{tabular*}{\textwidth}{@{\extracolsep{\fill}}lccccccccc@{\extracolsep{\fill}}}
\toprule%
& \multicolumn{2}{@{}c@{}}{Accuracy(\%)$\uparrow$} & \multicolumn{2}{@{}c@{}}{Precision(\%)$\uparrow$} &  \multicolumn{2}{@{}c@{}}{FPR(\%)$\downarrow$} & \multicolumn{2}{@{}c@{}}{FNR(\%)$\downarrow$} \\\cmidrule{2-3}\cmidrule{4-5}\cmidrule{6-7}\cmidrule{8-9}%
Dataset (Classifier) & STD & ZFN & STD & ZFN & STD & ZFN & STD & ZFN
\\\cmidrule{2-2}\cmidrule{3-3}\cmidrule{4-4}\cmidrule{5-5}\cmidrule{6-6}\cmidrule{7-7}\cmidrule{8-8}\cmidrule{9-9}%
PCBA (ET) &
\hspace{0.3cm}95.69\hspace{0.3cm} & 87.93 &
\hspace{0.3cm}98.18\hspace{0.3cm} & 80.56 & 
\hspace{0.3cm}1.72\hspace{0.3cm} & 24.14 & 
\hspace{0.3cm}6.9\hspace{0.3cm} & 0 \\
Cable (XGBoost) &
76.82 & 57.94 &
80.65 & 48.42 & 
8.51 & 69.5 & 
45.65 & 0 \\
Carpet (LR) &
85.6 & 50.21 &
80 & 42.38 & 
11.69 & 78.57 & 
19.1 & 0 \\
Grid (LR) &
95.98 & 85.43 &
94.55 & 66.28 & 
2.11 & 20.42 & 
8.77 & 0 \\
Hazelnut (LGBM) &
98.95 & 98.25 &
98.55 & 93.33 & 
0.47 & 2.33 & 
2.86 & 0 \\
Leather (XGBoost) &
92.17 & 90.43 &
91.11 & 80.7 & 
5.8 & 15.94 & 
10.87 & 0 \\
Screw (ADA) &
93 & 83.67 &
92.24 & 70.83 & 
4.97 & 27.07 & 
10.08 & 0 \\
Transistor (LGBM) &
88.7 & 49.15 &
81.25 & 30.77 & 
4.38 & 65.69 & 
35 & 0 \\
Zipper (LGBM) &
92.55 & 81.57 &
93.1 & 71.69 & 
5.88 & 34.56 & 
9.24 & 0 \\
\botrule
\end{tabular*}}
\end{minipage}
\end{center}
\end{table}
This table shows that, under the ZFN constraint, datasets like Hazelnut, Leather, Grid or Screw show relatively low false-positive rates (2.33\%, 15.94\%, 20.42\% and 27.07\% respectively), thanks to a correct anomalies reconstruction and a clear distinction between small normal and abnormal variations, inherent to the image complexity. Unlike these images, Figure~\ref{fig:Quanti1} shows how the reconstruction model has difficulties in following with fidelity the data distribution for the Carpet dataset, or recovering the large defects of the Cable dataset. 
In these cases, the patching technique cannot efficiently focus on the anomalies, and is completely fooled by the small normal variations. 

These observations prove that the reconstruction quality, the data distribution fidelity, and the weighting mask are crucial elements for the anomaly detection task under the zero-false-negative constraint.
If we do not consider the ZFN constraint, we can see that the method generates many more false negatives than false positives for all the datasets. This proves the difficulty of capturing the anomaly features in the entire overcrowded information contained in each image. 

\section{Conclusion}\label{sec7}
An anomaly detection methodology suited for a real-world industrial use case is developed in this work. This is the continuation of a first work leading the \textit{GanoDIP} method. The poor number of abnormal images (18) was the main limitation, thus yielding implementation difficulties due to the lack of defect variability. The current dataset contains around 10 times more abnormal images, with larger defect size, structure, and area variability. This amount is better, although still very small compared to the majority class. The proposal of this work is, therefore, to (i) be able to localize small defects, (ii) satisfy the zero-false-negative constraint, and (iii) reach the lowest false-positives rate possible. The \textit{VQGanoDIP} methodology detailed in this paper reached the objectives thanks to a three-step methodology. It takes advantage of the vast amount of normal data, instead of a regular binary classifier. After a reduced anomalies estimation technique, the few abnormal data are indeed kept for further less-data-intensive processing.

The first step takes advantage of the recent advances in terms of image synthesis that make it possible to reconstruct an original image, following an input data distribution being anomaly-free. The \textit{VQGAN} method, placed in an anomaly detection architecture, yields a strong representation of the normal class. It reconstructs very similar images to the original ones and, if any, replaces an abnormal set of pixels with an estimated normal one. The technique allows high-resolution reconstruction, fulfilling the business case constraint, requiring to deal with small defects like solder bridges or electronic component shifts. Furthermore, only the majority class is required to train the reconstruction model in an unsupervised manner. 
Therefore, the imbalanced learning specificity is managed at this stage, and we save the few minority-class images that we have for the next step and the test set.

The second step of the methodology is a comparison of the $1024 \times 1024$ original and reconstructed images. A significant number of appropriate metrics are extracted from this comparison, including the different neural networks losses, as well as the computer vision distances on the worst difference patches, following the \textit{GanoDIP} method strategy. To do so, a zoom-out-and-shift technique is performed on the worst patches to focus the metrics extraction on the highest Frechet Inception Distance areas. The objective is to make decisions on perceptual difference meanings instead of a regular absolute pixel difference. This step is applied to a balanced number of normal and abnormal class images, this number being conditioned by the abnormal set of images at hand. 

The last step is to train a classifier able to act as the anomaly score to determine the image class. Its goal is to discriminate between normal and abnormal images, thanks to the metrics collected previously. The zero-false-negative constraint requires to set a low probability threshold on the classifier prediction, generating more false positives as a regular accuracy setup. The price to ensure the quality requirements is an accuracy decrease by 7.76\%, reaching \textbf{87.93}\% (instead of 95.69\%). 

For the business use case, the proposed methodology achieves a drop of the current inspection time \textbf{from 8 seconds to 1.9 second}, and an estimated operator misjudgment rate \textbf{divided by 4}, which is a very satisfying achievement. The methodology developed can be used as a baseline for many other use cases, where high-resolution images with small details and low variation between normal and abnormal areas are considered, especially under the zero-false-negative constraint.

These promising results open new research questions for future works. Specifically, it would be interesting to find a better strategy to let the classifier learn the decision function integrating the zero-false-negative constraint directly, in an end-to-end manner. Another research direction could be to develop a reconstruction model that performs well on any type of dataset, whatever the normal variations characteristic or the anomaly sizes.

\bmhead{Acknowledgments}

The authors thank Jérôme Fink and Géraldin Nanfack for their insightful comments and discussions on this paper.

\section*{Declarations}


\begin{itemize}
\item Funding: Not Applicable
\item Conflict of interest/Competing interests: The authors declare that they have no competing or conflict of interests.
\item Ethics approval: The authors declare that this work is original, is not under consideration for publication elsewhere and has not been published previously. The authors approve the manuscript enclosed.
\item Consent to participate: Not Applicable
\item Consent for publication: The authors consent to the publication of this work.
\item Availability of data and materials: All public works and datasets have been cited in reference.
\item Code availability: Not Applicable
\item Authors' contributions: Arnaud Bougaham conceptualized the ideas, designed the algorithm, carried out the experiments and wrote the manuscript. Mohammed El Adoui, Isabelle Linden and Benoît Frénay supervised the work, provided critical improvements, helped writing, reviewed and approved the manuscript.\end{itemize}
\noindent

\bibliography{sn-bibliography}


\end{document}